\PassOptionsToPackage{table}{xcolor}
\documentclass{article}
\usepackage{arxiv}

\usepackage[utf8]{inputenc} 
\usepackage[T1]{fontenc}    
\usepackage[breaklinks]{hyperref} 
\usepackage{url}            
\usepackage{booktabs}       
\usepackage{amsfonts}       
\usepackage{nicefrac}       
\usepackage{microtype}      
\usepackage{graphicx}
\usepackage{natbib}
\usepackage{doi}
\usepackage{amsmath}
\usepackage{euscript}
\usepackage{multirow}
\usepackage{subcaption}
\usepackage[table]{xcolor}
\usepackage{tikz}
\usetikzlibrary{arrows.meta}
\usepackage{listings}
\usepackage{verbatim}
\usepackage{dirtytalk}
\usepackage{nicematrix}
\NiceMatrixOptions{
code-for-first-row = \color{gray} ,
code-for-first-col = \color{gray} ,
}
\usepackage{adjustbox} 
\usepackage{pifont}
\newcommand{\cmark}{\ding{51}}
\newcommand{\xmark}{\ding{55}}
\usepackage{makecell}
 \usepackage{relsize}
 \usepackage{pdflscape}
\usepackage{afterpage}
\usepackage{cprotect}

\definecolor{tabcolor1}{RGB}{223,194,125} 
\definecolor{tabcolor2}{RGB}{1,133,113} 
\definecolor{tabcolor3}{RGB}{166,97,26} 

\usepackage{listings}
\definecolor{gray}{rgb}{0.5,0.5,0.5}
\definecolor{gray2}{rgb}{0.6,0.6,0.6}
\title{Metrics Matter in Surgical Phase Recognition}

\date{} 					

\author{
    \href{https://orcid.org/0000-0002-4566-3049}{\includegraphics[scale=0.06]{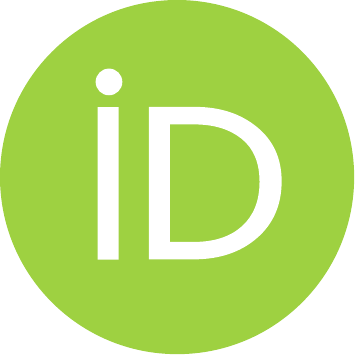}\hspace{1mm}
    Isabel Funke} \\ 
    Translational Surgical Oncology \\
    National Center for Tumor Diseases (NCT) \\
    \emph{and} Centre for Tactile Internet \\ 
    with Human-in-the-Loop (CeTI) \\
    Dresden, Germany \\
    \texttt{isabel.funke@nct-dresden.de} \\
    \And
    Dominik Rivoir \\ 
    Translational Surgical Oncology \\
    National Center for Tumor Diseases (NCT) \\
    \emph{and} Centre for Tactile Internet \\
    with Human-in-the-Loop (CeTI) \\
    Dresden, Germany \\
    \texttt{dominik.rivoir@nct-dresden.de} \\
    \And
    Stefanie Speidel \\ 
    Translational Surgical Oncology \\
    National Center for Tumor Diseases (NCT) \\
    \emph{and} Centre for Tactile Internet \\
    with Human-in-the-Loop (CeTI) \\
    Dresden, Germany \\
    \texttt{stefanie.speidel@nct-dresden.de} \\
}

\renewcommand{\headeright}{Technical Report}
\renewcommand{\undertitle}{Technical Report}

\begin{document}
\maketitle

\begin{abstract}
Surgical phase recognition is a basic component for different context-aware applications in computer- and robot-assisted surgery. In recent years, several methods for automatic surgical phase recognition have been proposed, showing promising results. 
However, a meaningful comparison of these methods is difficult due to differences in the evaluation process and incomplete reporting of evaluation details. In particular, the details of metric computation can vary widely between different studies. 
To raise awareness of potential inconsistencies, this paper summarizes common deviations in the evaluation of phase recognition algorithms on the Cholec80 benchmark. 
In addition, a structured overview of previously reported evaluation results on Cholec80 is provided, taking known differences in evaluation protocols into account. 
Greater attention to evaluation details could help achieve more consistent and comparable results on the surgical phase recognition task, leading to more reliable conclusions about advancements in the field and, finally, translation into clinical practice.

\end{abstract}

\keywords{Surgical phase recognition \and Benchmarking \and Laparoscopic video \and Cholec80}




\section{Introduction}

A first step towards introducing autonomous context-aware computer assistance to the operating room is to recognize automatically which surgical step, or phase, is being performed by the medical team. 
Therefore, the problem of \emph{surgical phase recognition} has received more and more attention in the research community.
Here, \emph{Cholec80} \citep{twinanda2016endonet}, a relatively large, annotated, public data set of cholecystectomy procedures, has become a popular benchmark for evaluating different methods and, thus, measuring progress in the field. 

However, a meaningful comparison of different methods is difficult since the evaluation protocols can vary substantially, starting with the division of Cholec80~data into training, validation, and test set.
In addition, the definitions of common evaluation metrics for surgical phase recognition leave some room for interpretation, meaning that different authors may compute evaluation metrics in different ways.
For example, it became popular to compute metrics with \say{relaxed boundaries}, meaning that errors around phase transitions are ignored under certain conditions.
Still, many researchers seem not to be aware of potential differences in the evaluation protocol when comparing their method to prior art. Yet, comparisons of incompatible results, see for example Fig.~\ref{fig:teaser}, are not conclusive.

In this paper, we summarize common deviations in the evaluation of surgical phase recognition algorithms on the Cholec80 benchmark.
We present how evaluation metrics are computed and how different implementations may differ. 
To further draw attention to the effects that evaluation details may have on the final results,
 we comprehensively evaluate a baseline model for phase recognition using different variants of common evaluation metrics.\footnote{The evaluation code is available at \url{https://gitlab.com/nct_tso_public/phasemetrics}.}

Finally, we review recent methods for surgical phase recognition and summarize the reported evaluation results on Cholec80. While one of our aims is to give an overview of the current state of the art in the field, it is almost impossible to draw final conclusions due to different -- and often also unclear -- evaluation procedures.
Still, we hope that our efforts 
to present previously published results in a structured and consistent way will make it easier to position future work with respect to the state of the art.

\newcommand*\annotatedFigureBoxCustom[8]{\draw[#5,thick] (#1) rectangle (#2);\node at (#4) [fill=#6,thick,shape=circle,draw=#7,inner sep=1pt,font=\tiny\sffamily,text=#8] {\textbf{#3}};}
\newcommand*\annotatedFigureBox[5]{\annotatedFigureBoxCustom{#1}{#2}{#3}{#4}{#5}{white}{#5}{#5}}
\newcommand*\annotatedFigureText[3]{\node at (#1) [fill=white,thick,shape=circle,draw=#2,inner sep=1pt,font=\tiny\sffamily,text=#2] {\textbf{#3}};}
\newenvironment {annotatedFigure}[1]{\centering\begin{tikzpicture}
\node[anchor=south west,inner sep=0] (image) at (0,0) { #1};\begin{scope}[x={(image.south east)},y={(image.north west)}]}{\end{scope}\end{tikzpicture}}

\begin{figure}[tb]
\begin{adjustbox}{width={0.9\textwidth},keepaspectratio,center}
\begin{annotatedFigure}
	{\includegraphics[width=1.0\linewidth]{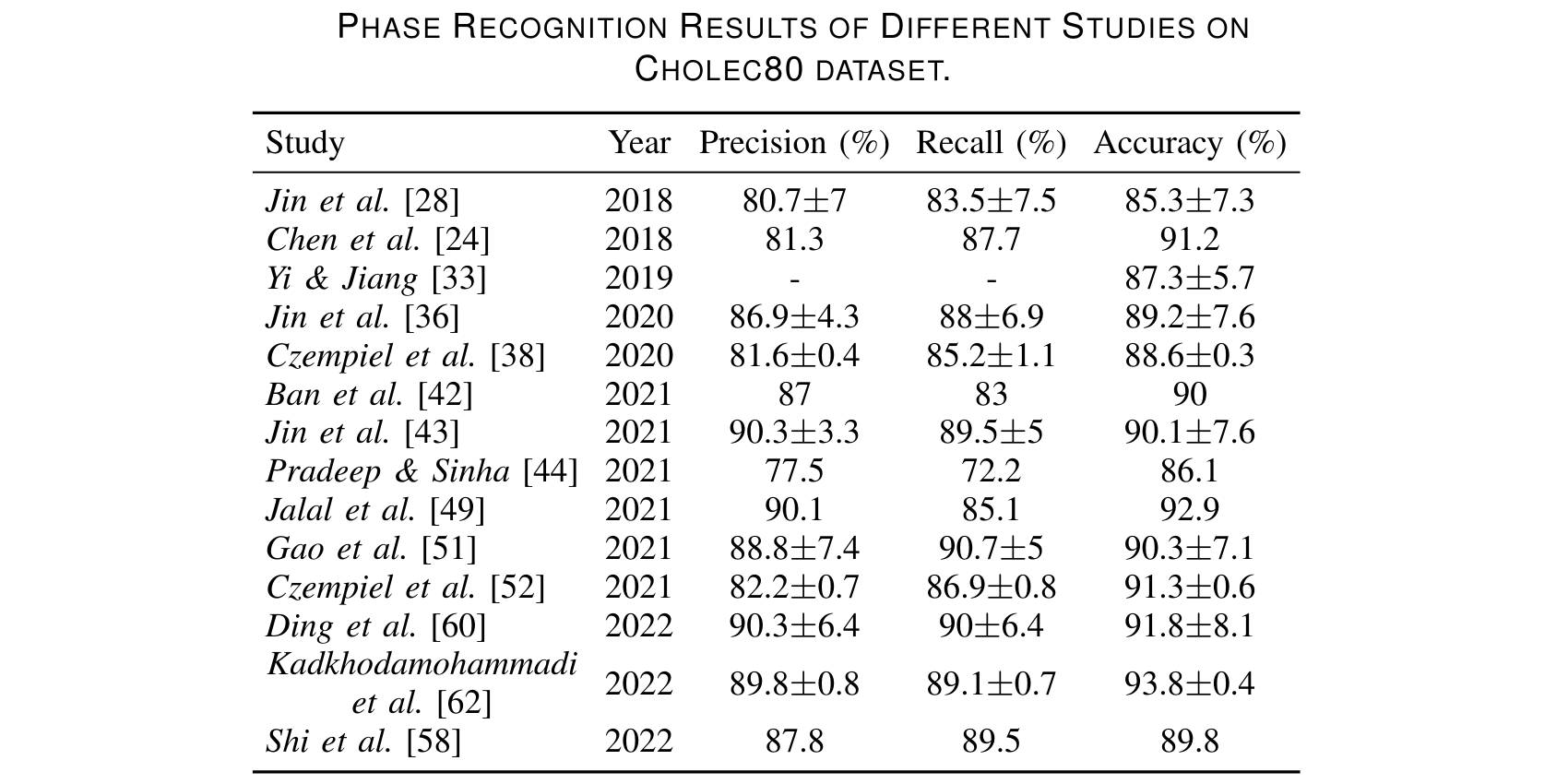}}
	\annotatedFigureBox{0.458,0.722}{0.811,0.772}{A}{0.458,0.722}{tabcolor1}
    \annotatedFigureBox{0.458,0.625}{0.811,0.672}{A}{0.458,0.622}{tabcolor1}
    \annotatedFigureBox{0.458,0.572}{0.811,0.617}{A}{0.458,0.572}{tabcolor1}
    \annotatedFigureBox{0.458,0.477}{0.811,0.521}{A}{0.458,0.477}{tabcolor1}
	\annotatedFigureBox{0.458,0.422}{0.811,0.470}{A}{0.458,0.422}{tabcolor1}
	\annotatedFigureBox{0.458,0.275}{0.811,0.322}{A}{0.458,0.275}{tabcolor1}
	\annotatedFigureBox{0.458,0.172}{0.811,0.222}{A}{0.458,0.172}{tabcolor1}
	\annotatedFigureBox{0.521,0.522}{0.551,0.567}{B}{0.521,0.522}{tabcolor2}
	\annotatedFigureBox{0.645,0.522}{0.675,0.567}{B}{0.645,0.522}{tabcolor2}
	\annotatedFigureBox{0.773,0.522}{0.803,0.567}{B}{0.773,0.522}{tabcolor2}
	\annotatedFigureBox{0.520,0.228}{0.550,0.272}{B}{0.520,0.225}{tabcolor2}
	\annotatedFigureBox{0.645,0.228}{0.675,0.272}{B}{0.645,0.225}{tabcolor2}
	\annotatedFigureBox{0.773,0.228}{0.803,0.272}{B}{0.773,0.225}{tabcolor2}
	\annotatedFigureBox{0.520,0.105}{0.550,0.150}{B}{0.520,0.105}{tabcolor2}
	\annotatedFigureBox{0.645,0.105}{0.675,0.150}{B}{0.645,0.105}{tabcolor2}
	\annotatedFigureBox{0.773,0.105}{0.803,0.150}{B}{0.773,0.105}{tabcolor2}
	\annotatedFigureText{0.82,0.695}{tabcolor3}{C1}
	\annotatedFigureText{0.82,0.545}{tabcolor3}{C2}
	\annotatedFigureText{0.82,0.247}{tabcolor3}{C3}
	\annotatedFigureText{0.82,0.126}{tabcolor3}{C4}
\end{annotatedFigure}
\end{adjustbox}
\caption{Phase recognition results on Cholec80 are often not comparable because some studies (A)~calculate metrics with \hyperref[subsec:relaxed]{relaxed boundaries}, (B)~calculate the standard deviation over experiment repetitions instead of over video-wise results, or (C)~deviate from the common 32:8:40 data split and use (1)~4-fold cross-validation on the test data, (2)~a 40:8:32 data split, (3)~a 48:12:20 data split, or (4)~a split with 60 videos for training and 20 for testing. The table is taken from a recent preprint on surgical phase recognition by \citet{demirdeep}.}
\label{fig:teaser}
\end{figure}
 
\section{Background}

\subsection{Surgical phase recognition}

Surgical phase recognition is the task to automatically estimate which phase is being performed by the medical team at each time during a surgical intervention. 
Since live video footage of the surgical field is available during many types of interventions,
 it is common to utilize the video data as primary source of information. 
In this case, given the set of surgical phases~$P$ and a video~$v$ of length~$T$, the task is to estimate which phase $y(v)_t \in P$ is being executed at each time $t$ in the video, where $0 \leq t < T$.

Surgical phase recognition can either be performed live, \emph{i.e.},
 online, or post-operatively, \emph{i.e.}, offline. For \emph{online} recognition, only information from video frames at previous times $t' \leq t$ can be utilized to estimate the phase at current time $t$. For \emph{offline} recognition, however, information from all frames in the video can be accessed.

The ultimate goal of automatic surgical phase recognition is to increase patient safety and resource efficiency.
In online mode, surgical phase recognition is an important component for automatic monitoring of surgical interventions and context-aware computer assistance.
In offline mode, surgical phase recognition can be useful for documentation purposes and content-based search of surgical video databases.

\pagebreak
\paragraph{Formal notation}
Let $P$ denote the set of surgical phases. The video $v$ is a sequence of video frames $v_t$, where $t$ refers to discrete time steps, $0 \leq t < T$. 
The sequence of known true phase labels in $v$ is $y(v) := \left(y(v)_t\right)_{0 \leq t < T}$ and the sequence of predicted phase labels, computed by a method for automatic phase recognition, is $\hat{y}(v) := \left(\hat{y}(v)_t\right)_{0 \leq t < T}$\,, where $y(v)_t \in P$ and $\hat{y}(v)_t \in P$. 
The temporal distance $\Delta t$ between two consecutive time steps corresponds to the temporal resolution at which the video is analyzed. In most cases, $\Delta t = 1\,\mathrm{s}$.

\subsection{Cholec80 data set}

\emph{Cholec80}\footnote{\url{http://camma.u-strasbg.fr/datasets}}~\citep{twinanda2016endonet} was one of the first public data sets for video-based surgical phase recognition and surgical tool recognition. Its release paved the way for recent progress and technical innovation in the field of surgical workflow analysis, where Cholec80 serves as a popular benchmark for comparing different algorithms. 

\begin{figure}[tb]
     \centering
     \begin{subfigure}[b]{0.35\textwidth}         
         \centering
        \begin{tabular}{ll}
            \toprule
            $p$ & Phase \\
            \midrule
            0 & Preparation \\
            1 & Calot triangle dissection \\
            2 & Clipping and cutting \\
            3 & Gallbladder dissection \\
            4 & Gallbladder packaging \\
            5 & Cleaning and coagulation \\
            6 & Gallbladder retraction \\
            \bottomrule
        \end{tabular}
        \caption{Surgical phases in Cholec80.}
        \label{tab:phases}
     \end{subfigure}
     \hfill
     \begin{subfigure}[b]{0.6\textwidth}
        \centering
        \begin{tikzpicture}
        \begin{scope}[every node/.style={circle,thick,draw}]
            \node (0) at (0.5,0) {0};
            \node (1) at (2,0) {1};
            \node (2) at (3.5,0) {2};
            \node (3) at (5,0) {3};
            \node (4) at (6.5,1) {4};
            \node (5) at (6.5,-1) {5};
            \node (6) at (8,0) {6};
        \end{scope}
        \begin{scope}[>={latex}, every edge/.style={draw=black, thick}]
            \path [->] (0) edge node {} (1);
            \path [->] (1) edge node {} (2);
            \path [->] (2) edge node {} (3);
            \path [->] (3) edge node {} (4);
            \path [->] (3) edge node {} (5);
            \path [->] (4) edge[bend right=15] node {} (5);
            \path [->] (4) edge node {} (6);
            \path [->] (5) edge[bend right=15] node {} (4); 
            \path [->] (5) edge[bend right=15] node {} (6); 
            \path [->] (6) edge[bend right=15] node {} (5);  
        \end{scope}
        \end{tikzpicture}    
        \caption{Cholec80 workflow modeled as a graph $(P, \EuScript{T})$ to show in which order surgical phases can appear. 
        The set of edges $\EuScript{T} \subset P \times P$ consists of those pairs~$(p, q)$ for which it is true that phase $q$ can \emph{immediately follow} phase $p$.} 
        \label{fig:phase-order}
     \end{subfigure}
     \hfill
     \caption{The Cholec80 workflow.}
\end{figure}
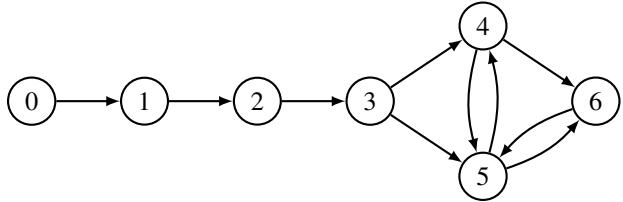

Cholec80 consists of 80 video recordings of laparoscopic cholecystectomies, \emph{i.e.}, minimally invasive surgeries to remove the gallbladder.
The pre-defined set $P$ of cholecystectomy phases consists of seven phases, see Table~\ref{tab:phases}.
We usually refer to a surgical phase in $P$ by means of its identifier $p$, with $0 \leq p < \vert P \vert$.
All Cholec80 videos are annotated with phase information and information regarding the presence of seven distinct surgical tools. 
There are known constraints on the order of phases in Cholec80: If phase~$p_i$ is immediately followed by phase~$p_{i+1}$, then it must hold that ($p_i, p_{i + 1}) \in \EuScript{T}$, where $\EuScript{T}$ denotes the set of valid phase transitions, see Fig.~\ref{fig:phase-order}.

In Cholec80, most of the video frames belong to phases 1 or 3, which have a median duration of almost $14\,\mathrm{min}$ and almost $10\,\mathrm{min}$, respectively, per video. The remaining phases are underrepresented with a median duration between $1\,\mathrm{min}$ and a bit over $2\,\mathrm{min}$. Fig.~\ref{fig:phase-duration} presents the distributions of phase durations in more detail. Clearly, Cholec80 is an \emph{imbalanced} data set with majority classes phase~1 and phase~3.

\section{Evaluation on Cholec80}
\label{sec:eval}

\subsection{Data splits}
\label{sec:datasplits}
Benchmark data sets for evaluating machine learning algorithms are usually split into three disjoint subsets: (1) training data for model training, (2) validation data for tuning models during their development, and (3) test data for model evaluation on previously unseen data.

For the Cholec80 data set, \citet{twinanda2016endonet} initially proposed to use the first 40 videos for training the visual feature extractor. 
Then, they trained and tested the temporal model on the remaining 40 videos, using a 4-fold cross-validation setup. Here, each validation fold consisted of 10 videos and the training data consisted of the videos from the remaining folds, \emph{i.e.}, 30 videos. In following experiments, \citep[chapter 6.4]{twinanda2017vision} also included the first 40~videos for training the temporal model, thus training on 70 videos for each validation fold.

In contrast, the first model that was evaluated on Cholec80 after EndoNet, namely, SV-RCNet \citep{jin2017sv}, was trained end to end in one training step. Thus, the authors proposed to train the model on the first 40 videos in Cholec80 and test it on the remaining 40 videos -- skipping the cross-validation procedure. With the exception of Endo3D \citep{chen2018endo3d},
 most of the following studies adopted this simple split into training and test data. Methods that are trained in multiple steps used this train\,/\,test split as well, meaning that the temporal model is trained on the same example videos as the visual feature extractor. Notably, this can make training difficult if the visual feature extractor fits the training data already perfectly, \emph{cf.} \citet{yi2022not}.

Unfortunately, splitting a data set only into training and test data usually implies that the test data is used for validation during model development, meaning that the model is both tuned and evaluated on the test data. In this case, the evaluation can yield overly good results, which, however, are bad estimates of the true generalization error.
For this reason, researchers proposed to set eight videos of the training data aside as validation data for hyperparameter tuning and model selection. In some cases, the final model is still trained on all 40 training videos, but this is usually not specified clearly. Some papers also describe ambiguous data splits, such as using \say{40 videos for training, 8 videos for validation and the remaining 40 videos for testing} \citep{zhang2021real} (similarly in \citet{ding2022exploring}). 
In contrast, \citet{czempiel2020tecno} used 40 videos for training, 8~for validation, and 32 for testing. Similarly, \citet{zhang2022retrieval} and \citet{zhang2022surgical} used 40 videos for training, 20 for validation, and 20 for testing.

Recently, researchers proposed further data splits on Cholec80 with the motivation to have more example videos available for training and validation. \citet{czempiel2021opera} split the data into 60 videos for training and validation and 20 videos for testing. For hyperparameter and model selection, they performed 5-fold cross-validation on the 60 videos, thus training on 48 videos and validating on 12 videos in each fold. Finally, they evaluated all five models on the hold-out test set and reported the averaged results.
Similarly, \citet{kadkhodamohammadi2022patg} proposed to train on 60 videos and test on 20 videos. In addition, they repeated the experiment five times, where they randomly sampled 20 test videos from the original 40 test videos -- \emph{i.e.}, the ones proposed by \citet{jin2017sv} -- in each repetition.

\subsection{Evaluation metrics}
\label{sec:metrics}
Common metrics to assess the classification performance of a phase recognition method are accuracy, macro-averaged precision, and macro-averaged recall \citep{padoy2012statistical}. 
Here, precision checks if a phase is recognized erroneously (false positive prediction) while recall checks whether parts of a phase are missed (false negative predictions).
Phase-wise precision and recall scores are somewhat complementary to each other, meaning that a higher precision score can be traded for a lower recall score and vice versa.
To measure both how accurately and how comprehensively a phase is recognized, F1 score or Jaccard index can be used. 

The evaluation metrics are usually computed \emph{video-wise}, \emph{i.e.}, for each video individually, and then averaged over all videos in the test set. In the following, we define the video-wise evaluation metrics given prediction $\hat{y}(v)$ and ground truth video annotation $y(v)$.

For comparing $\hat{y}(v)$ against $y(v)$, the \emph{confusion matrix} $C\!\left(\hat{y}(v)\right) :=\bigl(C\!\left(\hat{y}(v)\right)_{pq}\bigr)_{0 \leq p, q < \vert P \vert} \in \mathbb{N}^{\vert P \vert \times \vert P \vert}$ is computed, where
$$C\!\left(\hat{y}(v)\right)_{pq}:= \lvert \{t : y(v)_t = p~\mathrm{and}~\hat{y}(v)_t = q, 0 \leq t < T\}\rvert \geq 0.$$ 
In other words, the entry in the $p$-th row and $q$-th column of the confusion matrix counts how many frames 
are annotated as phase $p$ and predicted as phase $q$. 

\paragraph{Note.} We denote the confusion matrix and the video-wise evaluation metrics in dependency of the prediction~$\hat{y}(v)$ and do not explicitly mention the annotation~$y(v)$, which is known and fixed for any video. For convenience, we may also omit~$\hat{y}(v)$, \emph{e.g.}, write $C_{pq}$ instead of $C\!\left(\hat{y}(v)\right)_{pq}$, when it is clear 
to which video prediction (and annotation) we refer. 

For phase $p \in P$, the numbers of true positive ($\mathsf{TP}_p$), false positive ($\mathsf{FP}_p$), and false negative ($\mathsf{FN}_p$) predictions are computed as:
\begin{align*}
    \mathsf{TP}_p\!\left(\hat{y}(v)\right) &:= \vert \{t : y(v)_t = p~\mathrm{and}~\hat{y}(v)_t = p, 0 \leq t < T\}\vert = C_{pp} \\
    \mathsf{FN}_p\!\left(\hat{y}(v)\right) &:= \vert \{t : y(v)_t = p~\mathrm{and}~\hat{y}(v)_t \neq p, 0 \leq t < T\}\vert = \sum_{q \neq p} C_{pq} \\
    \mathsf{FP}_p\!\left(\hat{y}(v)\right) &:= \vert \{t : y(v)_t \neq p~\mathrm{and}~\hat{y}(v)_t = p, 0 \leq t < T\}\vert = \sum_{q \neq p} C_{qp}
\end{align*}
Thus, $\mathsf{TP}_p$ counts how many frames of phase $p$ are correctly predicted as phase $p$, $ \mathsf{FN}_p$ counts how many frames of phase $p$ are incorrectly predicted as another phase $q \neq p$, and $\mathsf{FP}_p$ counts how many frames of other phases $q \neq p$ are incorrectly predicted as phase $p$. 

\emph{Phase-wise} video-wise evaluation metrics are defined as: 
\begin{align*}
    \mathsf{Precision}_p\!\left(\hat{y}(v)\right) &:= \frac{\vert \{t : y(v)_t = p~\mathrm{and}~\hat{y}(v)_t = p, 0 \leq t < T\}\vert}{\vert \{t : \hat{y}(v)_t = p, 0 \leq t < T\}\vert} = \frac{C_{pp}}{\sum_q{C_{qp}}} = \frac{\mathsf{TP}_p}{\mathsf{TP}_p + \mathsf{FP}_p} \\
    \mathsf{Recall}_p\!\left(\hat{y}(v)\right) &:= \frac{\vert \{t : y(v)_t = p~\mathrm{and}~\hat{y}(v)_t = p, 0 \leq t < T\}\vert}{\vert \{t : y(v)_t = p, 0 \leq t < T\}\vert} = \frac{C_{pp}}{\sum_q{C_{pq}}} = \frac{\mathsf{TP}_p}{\mathsf{TP}_p + \mathsf{FN}_p} \\
    \mathsf{F1}_p\!\left(\hat{y}(v)\right) &:= \frac{2 \cdot \mathsf{Precision}_p\!\left(\hat{y}(v)\right) \cdot \mathsf{Recall}_p\!\left(\hat{y}(v)\right)}{\mathsf{Precision}_p\!\left(\hat{y}(v)\right) + \mathsf{Recall}_p\!\left(\hat{y}(v)\right)} = \frac{2 \cdot C_{pp}}{\sum_q{C_{pq}} + \sum_q{C_{qp}}} = \frac{2 \cdot \mathsf{TP}_p}{2 \cdot \mathsf{TP}_p + \mathsf{FP}_p + \mathsf{FN}_p} \\
    \begin{split}
    \mathsf{Jaccard}_p\!\left(\hat{y}(v)\right) &:= \frac{\vert S_p \cap \hat{S}_p \vert}{\vert S_p \cup \hat{S}_p \vert}, \mathrm{where~} S_p := \{t : y(v)_t = p, 0 \leq t < T\} \mathrm{~and~} \hat{S}_p := \{t : \hat{y}(v)_t = p, 0 \leq t < T\} \\    
    &= \frac{\vert S_p \cap \hat{S}_p \vert}{\vert S_p \vert + \vert \hat{S}_p \vert - \vert S_p \cap \hat{S}_p \vert} = \frac{C_{pp}}{\sum_q{C_{pq}} + \sum_q{C_{qp}} - C_{pp}} = \frac{\mathsf{TP}_p}{\mathsf{TP}_p + \mathsf{FP}_p + \mathsf{FN}_p}
    \end{split}
\end{align*}
$\mathsf{Precision}_p$ describes how many frames that are predicted as phase $p$ are actually annotated as phase $p$ and therefore penalizes false positive predictions. In contrast, $\mathsf{Recall}_p$ describes how many frames of phase $p$ are actually predicted as phase $p$ and therefore penalizes false negative predictions. To consider both true negative and false negative predictions, $\mathsf{F1}_p$ (also known as Dice similarity coefficient) is defined as the harmonic mean of $\mathsf{Precision}_p$ and $\mathsf{Recall}_p$. 
$\mathsf{Jaccard}_p$ is defined as the \emph{intersection over union} of the set $S_p$, which refers to the video frames that are annotated as phase $p$, and the set $\hat{S}_p$, which refers to the video frames that are predicted as phase $p$. As shown above, $\mathsf{Jaccard}_p$ is calculated similarly to $\mathsf{F1}_p$ but does not count the true positives twice. 

The overall \emph{macro-averaged} video-wise metrics are obtained by computing the average over all phases:
$$ \mathsf{Macro\,Precision}\!\left(\hat{y}(v)\right) := \frac{1}{\vert P \vert} \sum_{p \in P}{\mathsf{Precision}_p}\!\left(\hat{y}(v)\right),~ \mathsf{Macro\,Recall}\!\left(\hat{y}(v)\right) := \frac{1}{\vert P \vert} \sum_{p \in P}{\mathsf{Recall}_p}\!\left(\hat{y}(v)\right),$$
$$\mathsf{Macro\,F1}\!\left(\hat{y}(v)\right) := \frac{1}{\vert P \vert} \sum_{p \in P}{\mathsf{F1}_p}\!\left(\hat{y}(v)\right),~ \mathsf{Macro\,Jaccard}\!\left(\hat{y}(v)\right) := \frac{1}{\vert P \vert} \sum_{p \in P}{\mathsf{Jaccard}_p}\!\left(\hat{y}(v)\right)$$
In addition, the accuracy metric is defined as the fraction of correct predictions in the video:
$$\mathsf{Accuracy}\!\left(\hat{y}(v)\right) := \frac{\vert \{t : y(v)_t = \hat{y}(v)_t, 0 \leq t < T\} \vert}{T} = \frac{\sum_p C_{pp}}{\sum_p \sum_q C_{pq}}$$
All evaluation metrics yield values between zero and one, where one indicates a perfect prediction. In many cases, authors report the numbers after multiplication with $100\,\%$.

Given a test set $V$ of videos $v$ and the set of predictions $\left\{ \hat{y}(v) : v \in V \right\}$, 
evaluation results are computed for each test video, yielding the set of evaluation results $\left\{ \mathsf{Metric}\!\left(\hat{y}(v)\right) : v \in V \right\}$.
Based on this, the sample mean, \emph{i.e.}, $\mathrm{mean}\left\{ \mathsf{Metric}\!\left(\hat{y}(v)\right) : v \in V \right\}$, and the corrected sample standard deviation, \emph{i.e.}, $\mathrm{std}\left\{ \mathsf{Metric}\!\left(\hat{y}(v)\right) : v \in V \right\}$, are computed over all videos\footnote{
As commonly known, given a set (or sample) $X$ of numbers $x \in X$,
\begin{align*}
    \mathrm{mean}\left\{ x : x \in X \right\} &:= \frac{1}{|X|} \sum_{x \in X} x \\
    \mathrm{std}\left\{ x : x \in X \right\} &:= \sqrt{\frac{1}{|X| - 1} \sum_{x \in X} \bigl( x -  \frac{1}{|X|} \sum_{x' \in X} x' \bigr)^2}
\end{align*}
}. Here, $\mathsf{Metric} \in \left\{ \mathsf{Accuracy}, \mathsf{Macro\,Precision}, \mathsf{Macro\,Recall}, \mathsf{Macro\,F1}, \mathsf{Macro\,Jaccard} \right\}$.

\paragraph{Phase-wise performance}
It can be insightful to examine how well a phase recognition method performs on each of the surgical phases. For phase $p \in P$ and $\mathsf{Metric} \in \left\{ \mathsf{Precision}, \mathsf{Recall}, \mathsf{F1}, \mathsf{Jaccard} \right\}$, the phase-wise evaluation results can be summarized using the phase-wise mean, \emph{i.e.}, $\mathrm{mean}\left\{ \mathsf{Metric}_p\!\left(\hat{y}(v)\right) : v \in V \right\}$, and the phase-wise corrected standard deviation, \emph{i.e.}, $\mathrm{std}\left\{ \mathsf{Metric}_p\!\left(\hat{y}(v)\right) : v \in V \right\}$.
Based on that, the mean over all phase-wise means, \emph{i.e.}, $\mathrm{mean}\left\{ \mathrm{mean}\left\{ \mathsf{Metric}_p\!\left(\hat{y}(v)\right) : v \in V \right\} : p \in P \right\}$, can be calculated to summarize performance over all phases\footnote{
In general, ${\mathrm{mean}\left\{ \mathrm{mean}\left\{ \mathsf{Metric}_p\!\left(\hat{y}(v)\right) : v \in V \right\} : p \in P \right\} = \mathrm{mean}\left\{ \mathsf{Macro\,Metric}\!\left(\hat{y}(v)\right) : v \in V \right\}}$. 
}. 
In addition, the standard deviation over the phase-wise means, \emph{i.e.}, $\mathrm{std} \left\{\mathrm{mean}\left\{ \mathsf{Metric}_p\!\left(\hat{y}(v)\right) : v \in V \right\} : p \in P \right\}$ can be used to describe the variation of performance between different phases.

\paragraph{Summarizing results over several experimental runs}
Training a deep learning model involves several sources of randomness, including random shuffling of batches,
 random initialization of model components, data augmentation, and execution of optimized, non-deterministic routines on the GPU. Therefore, a single deep learning experiment, where a model is trained on the training data once and then evaluated on the test data, cannot be considered conclusive. In fact, good or bad evaluation results could be due to chance such as picking a good or bad random seed. 

For this reason, it is good practice to repeat each experiment $n$ times, using different random seeds, to obtain a more reliable estimate of model performance. In this case, we denote the prediction computed for test video $v \in V$ in the $i$-th experimental run as $\hat{y}_i(v)$, $0 \leq i < n$. 

For $\mathsf{Metric} \in \left\{ \mathsf{Precision}, \mathsf{Recall}, \mathsf{F1}, \mathsf{Jaccard} \right\}$, the set $\left\{ \mathsf{Metric}_p\!\left(\hat{y}_i(v)\right) : p\in P, v \in V,~0 \leq i < n \right\}$ of evaluation results can be summarized using the following metrics:
\begin{itemize}
    \item Overall mean: $M(\mathsf{Metric}) := \mathrm{mean} \left\{ \mathsf{Metric}_p\!\left(\hat{y}_i(v)\right) : p\in P, v \in V,~0 \leq i < n \right\}$ 
    \item Variation over videos: $SD_{V}(\mathsf{Metric}) := \mathrm{std} \left\{ \mathrm{mean} \left\{ \mathsf{Metric}_p\!\left(\hat{y}_i(v)\right) : p \in P,~0 \leq i < n \right\}: v \in V \right\}$
    \item Variation over phases: $SD_{P}(\mathsf{Metric}) := \mathrm{std} \left\{ \mathrm{mean} \left\{ \mathsf{Metric}_p\!\left(\hat{y}_i(v)\right) : v \in V,~0 \leq i < n \right\}: p \in P \right\}$
    \item Variation over runs: $SD_{R}(\mathsf{Metric}) := \mathrm{std} \left\{ \mathrm{mean} \left\{ \mathsf{Metric}_p\!\left(\hat{y}_i(v)\right) : p \in P, v \in V \right\}: 0 \leq i < n \right\}$
\end{itemize}

Similarly, $\mathsf{Accuracy}$, macro-averaged phase-wise metrics, \emph{e.g.}, $\mathsf{Macro\,F1}$\footnote{
Since $\mathsf{Macro\,F1}\!\left(\hat{y}_i(v)\right) = \mathrm{mean} \left\{ \mathsf{F1}_p\!\left(\hat{y}_i(v)\right) : p\in P \right\}$, it follows that $M(\mathsf{Macro\,F1}) = M(\mathsf{F1})$, $SD_V(\mathsf{Macro\,F1}) = SD_V(\mathsf{F1})$, and $SD_R(\mathsf{Macro\,F1}) = SD_R(\mathsf{F1})$. The same holds for the metrics $\mathsf{Precision}$, $\mathsf{Recall}$, and $\mathsf{Jaccard}$.
}, and metrics computed for individual phases, \emph{e.g.}, $\mathsf{F1}_p$, can be summarized using the mean, variation over videos, and variation over runs:
    \begin{itemize}
        \item $M(\mathsf{Metric}) := \mathrm{mean} \left\{ \mathsf{Metric}\!\left(\hat{y}_i(v)\right) : v \in V,~0 \leq i < n \right\}$
        \item  $SD_{V}(\mathsf{Metric}) := \mathrm{std} \left\{ \mathrm{mean} \left\{ \mathsf{Metric}\!\left(\hat{y}_i(v)\right) : 0 \leq i < n \right\} : v \in V \right\}$
        \item $SD_{R}(\mathsf{Metric}) := \mathrm{std} \left\{ \mathrm{mean} \left\{ \mathsf{Metric}\!\left(\hat{y}_i(v)\right) : v \in V \right\} : 0 \leq i < n \right\}$
    \end{itemize}

\subsection{Inconsistencies in calculating metrics}
\label{sec:metric-variants}
Unfortunately, authors usually refrain from specifying the details of how they calculated the evaluation results.
However, there are subtle differences and deviating definitions for computing the evaluation metrics, which could limit the comparability of evaluation results.
In the following, we present possible inconsistencies and different variants of the common evaluation metrics for surgical phase recognition.

\subsubsection{Calculating the standard deviation}
\label{sec:issue-std}
Authors rarely specify whether or not they apply Bessel's correction when calculating the standard deviation. Without Bessel's correction, they compute
$$\mathbf{std}~X := \sqrt{\frac{1}{|X|} \sum_{x \in X} \bigl( x -  \frac{1}{|X|} \sum_{x' \in X} x' \bigr)^2}$$
Clearly, $\mathbf{std}~X < \mathrm{std}~X$, especially in the case of small sample sizes $|X|$.
Some libraries, such as NumPy~\citep{harris2020array}, compute the uncorrected standard deviation by default. In contrast, MATLAB's \verb|std| function applies Bessel's correction by default. 

Furthermore, authors rarely specify explicitly over which source of variation the reported standard deviation was calculated. This can be confusing, such as in the case of relaxed metrics, see section~\ref{subsec:relaxed}, where the official evaluation script computes the standard deviation over videos for accuracy, but over phases for the remaining metrics.

\subsubsection{Calculating F1 scores}
\label{sec:issue-macro-f1}
An alternative definition of the macro-averaged video-wise F1 score is
\begin{equation*}
    \mathbf{Macro\,F1}\!\left(\hat{y}(v)\right) := \frac{2 \cdot \mathsf{Macro\,Precision}\!\left(\hat{y}(v)\right) \cdot \mathsf{Macro\,Recall}\!\left(\hat{y}(v)\right)}{\mathsf{Macro\,Precision}\!\left(\hat{y}(v)\right) + \mathsf{Macro\,Recall}\!\left(\hat{y}(v)\right)}
\end{equation*}
$\mathbf{Macro\,F1}$ is the harmonic mean computed over the arithmetic means over, respectively, $\mathsf{Precision}_p$ and $\mathsf{Recall}_p$, while $\mathsf{Macro\,F1}$ is the arithmetic mean computed over $\mathsf{F1}_p$, which in turn is the harmonic mean over $\mathsf{Precision}_p$ and $\mathsf{Recall}_p$.
These definitions are \emph{not} equivalent. In fact, \cite{opitz2019macro} show that $\mathbf{Macro\,F1} \geq \mathsf{Macro\,F1}$ and $\mathbf{Macro\,F1} > \mathsf{Macro\,F1}$ \emph{iff} $\mathsf{Precision}_p \neq \mathsf{Recall}_p$ for at least one phase $p \in P$. 
We argue that $\mathsf{Macro\,F1}$ is a more meaningful metric for surgical phase recognition because it measures how well, \emph{i.e.}, how accurately \emph{and} how comprehensively, each phase is recognized -- on average -- in a test video. In contrast, $\mathbf{Macro\,F1}$ balances two video-level scores, macro-averaged precision and  macro-averaged recall, by computing the harmonic mean.

The difference between $\mathbf{Macro\,F1}$ and $\mathsf{Macro\,F1}$ can be considerably large when  there  are  many  phases $p$ with ${\vert \mathsf{Precision}_p - \mathsf{Recall}_p}\vert \gg 0$.
To illustrate, \cite{opitz2019macro} construct the following example: Let $\vert P \vert$ be an even number and ${\mathsf{Precision}_p \rightarrow 1}$, $\mathsf{Recall}_p \rightarrow 0$ for half of the phases and, exactly the other way round, $\mathsf{Precision}_p \rightarrow 0$, $\mathsf{Recall}_p \rightarrow 1$ for the remaining phases. Then, $\mathsf{F1}_p \rightarrow 0$ for all phases and therefore $\mathsf{Macro\,F1} \rightarrow 0$. However, $\mathsf{Macro\,Precision} \approx \mathsf{Macro\,Recall} \approx 0.5$ and therefore $\mathbf{Macro\,F1} \approx 0.5 \gg \mathsf{Macro\,F1}$.

In some cases, authors simply report~$\mathit{F1}$, which is the harmonic mean of the overall mean precision and recall:
$$\mathit{F1} := \frac{2 \cdot M(\mathsf{Precision}) \cdot M(\mathsf{Recall})}{M(\mathsf{Precision}) + M(\mathsf{Recall})} = \frac{2 \cdot \frac{1}{|V|}\sum_{v \in V} \mathsf{Macro\,Precision}\!\left(\hat{y}(v)\right) \cdot \frac{1}{|V|}\sum_{v \in V} \mathsf{Macro\,Recall}\!\left(\hat{y}(v)\right)}{\frac{1}{|V|}\sum_{v \in V} \mathsf{Macro\,Precision}\!\left(\hat{y}(v)\right) + \frac{1}{|V|}\sum_{v \in V} \mathsf{Macro\,Recall}\!\left(\hat{y}(v)\right)}$$

However, following the same argument as above,
$$\mathit{F1} \geq M(\mathbf{Macro\,F1}) = \frac{1}{|V|}\sum_{v \in V} \mathbf{Macro\,F1}\!\left(\hat{y}(v)\right) = \frac{1}{|V|}\sum_{v \in V} \frac{2 \cdot \mathsf{Macro\,Precision}\!\left(\hat{y}(v)\right) \cdot \mathsf{Macro\,Recall}\!\left(\hat{y}(v)\right)}{\mathsf{Macro\,Precision}\!\left(\hat{y}(v)\right) + \mathsf{Macro\,Recall}\!\left(\hat{y}(v)\right)}$$
and $\mathit{F1} > M(\mathbf{Macro\,F1})$ \emph{iff} $\mathsf{Macro\,Precision}\!\left(\hat{y}(v)\right) \neq \mathsf{Macro\,Recall}\!\left(\hat{y}(v)\right)$ for at least one video $v \in V$. 

It follows that $\mathit{F1} \geq M(\mathbf{Macro\,F1}) \geq M(\mathsf{Macro\,F1})$. Thus, $\mathit{F1}$ is a -- not necessarily very tight -- upper bound for the actual number of interest, namely, $M(\mathsf{F1}) = M(\mathsf{Macro\,F1})$.

\subsubsection{\emph{Undefined} values when calculating phase-wise video-wise evaluation metrics}
\label{sec:undefined}
For any phase-wise metric, $\mathsf{Metric}_p\!\left(\hat{y}(v)\right)$ can be \emph{undefined} in the case that the denominator is zero. Table~\ref{tab:zerodiv} summarizes under which conditions this is the case: 
$\mathsf{Precision}_p$ is undefined when $\sum_q{C_{qp}} = 0$, \emph{i.e.}, none of the video frames is predicted as phase $p$. $\mathsf{Recall}_p$ is undefined when $\sum_q{C_{pq}} = 0$, \emph{i.e.}, none of the frames is annotated as phase $p$. $\mathsf{F1}_p$ and $\mathsf{Jaccard}_p$ are undefined when $\sum_q{C_{pq}} + \sum_q{C_{qp}} = 0$ (which also implies $C_{pp} = 0$), \emph{i.e.}, none of the frames is either annotated or predicted as phase $p$.

\begin{table}[htb]
	\caption{Results calculated by the phase-wise metrics in edge cases $(\sum_q{C_{pq}} = 0 \mathrm{~or~} \sum_q{C_{qp}} = 0)$.
    Undefined values, caused by division by zero, are denoted as n/a.
    When using $\mathsf{Strategy\,A}$ for handling undefined values, all \emph{undefined} values -- marked as n/a -- will be excluded from ensuing calculations.}
    \label{tab:zerodiv}
	\centering
	\begin{tabular}{lllllll}
		\toprule  
            &   & $\mathsf{Precision}_p\!\left(\hat{y}(v)\right)$   & $\mathsf{Recall}_p\!\left(\hat{y}(v)\right)$  & $\mathsf{F1}_p\!\left(\hat{y}(v)\right)$  & $\mathsf{Jaccard}_p\!\left(\hat{y}(v)\right)$ \\
        \cmidrule(r){3-6}
        $p \in y(v)$?  & $p \in \hat{y}(v)$?    &   &   &   &   \\  
		\cmidrule(r){1-2}
        no $\left(\sum_q{C_{pq}} = 0\right)$   & no $\left(\sum_q{C_{qp}} = 0\right)$  & n/a   & n/a   & n/a   & n/a   \\
                                    & yes $\left(\sum_q{C_{qp}} > 0\right)$ & $0$   & n/a   & $0$   & $0$ \\
        \cmidrule(r){1-2}
        yes $\left(\sum_q{C_{pq}} > 0\right)$  & no $\left(\sum_q{C_{qp}} = 0\right)$  & n/a   & $0$   & $0$   & $0$ \\
                                    & yes $\left(\sum_q{C_{qp}} > 0\right)$ & $\geq 0$ & $\geq 0$  & $\geq 0$  & $\geq 0$ \\
		\bottomrule
	\end{tabular}
\end{table}

In the case of Cholec80, specifically phases 0 and 5 are missing in some of the video annotations, making the occurrence of \emph{undefined} values unavoidable when calculating the video-wise evaluation metrics. 
Note that subsequent computations, such as computing macro-averaged metrics or the statistics $M$, $SD_V$, $SD_R$ of a phase-wise metric, are dependent on the values $\left\{ \mathsf{Metric}_p\!\left(\hat{y}(v)\right) : v \in V \right\}$.
Thus, it is desirable to handle undefined values in such a way that ensuing calculations to obtain summary metrics can still be performed.

There are different strategies for handling undefined values, and these strategies are rarely described explicitly. The popular \texttt{scikit-learn} library~\citep{scikit-learn} by default sets undefined values to zero, but offers an option to set them to one as well. In the first case, derived metrics can be unexpectedly low due to undefined values, in the latter case, derived metrics can be unexpectedly high. 
Another strategy ($\mathsf{Strategy~A}$, see Table~\ref{tab:zerodiv}) is to simply exclude any undefined values from ensuing calculations.\footnote{For example, $\mathsf{Strategy~A}$ was used by \cite{czempiel2020tecno} to compute precision and recall scores.} However, this can be overly restrictive for $\mathsf{Metric} \in \{\mathsf{Precision}, \mathsf{F1}, \mathsf{Jaccard}\}$. Here, $\mathsf{Metric}_p\!\left(\hat{y}(v)\right)$ will be ignored when phase $p$ is missing in the annotation and, \emph{correctly}, also in the prediction. On the other hand, $\mathsf{Metric}_p\!\left(\hat{y}(v)\right)$ will be zero as soon as there is a single false positive prediction of phase $p$. 
Consequently, one common strategy ($\mathsf{Strategy~B}$, see Table~\ref{tab:handleA}) is to exclude \emph{all} results $\left\{ \mathsf{Metric}_p\!\left(\hat{y}(v)\right) : \mathsf{Metric} \in \{ \mathsf{Precision}, \mathsf{Recall}, \mathsf{F1}, \mathsf{Jaccard} \} \right\}$ for a video $v$ where phase~$p$ is missing in the annotation, even if -- in the case of false positive predictions -- some of these values may be zero and \emph{not} undefined.\footnote{For example, $\mathsf{Strategy~B}$ was implemented for the relaxed metrics, see section~\ref{subsec:relaxed}.} 


\begin{table}[t!]
	\caption{$\mathsf{Strategy~B}$ for handling undefined values. If phase $p$ is missing in the annotation of video $v$, all corresponding phase-wise results 
    are excluded from ensuing calculations. Further \emph{undefined} values are excluded as well.}
	\centering
	\begin{tabular}{llllll}
		\toprule
            &   & $\mathsf{Precision}_p\!\left(\hat{y}(v)\right)$   & $\mathsf{Recall}_p\!\left(\hat{y}(v)\right)$  & $\mathsf{F1}_p\!\left(\hat{y}(v)\right)$  & $\mathsf{Jaccard}_p\!\left(\hat{y}(v)\right)$ \\
        \cmidrule(r){3-6}
        $p \in y(v)$?  & $p \in \hat{y}(v)$?    &   &   &   &   \\  
		\cmidrule(r){1-2}
        no $\left(\sum_q{C_{pq}} = 0\right)$   & no $\left(\sum_q{C_{qp}} = 0\right)$  & exclude   & exclude    & exclude    & exclude \\
                                    & yes $\left(\sum_q{C_{qp}} > 0\right)$ & exclude   & exclude    & exclude    & exclude \\
        \cmidrule(r){1-2}
        yes $\left(\sum_q{C_{pq}} > 0\right)$  & no $\left(\sum_q{C_{qp}} = 0\right)$  & exclude   & $0$       & $0$       & $0$ \\
		\bottomrule
	\end{tabular}
	\label{tab:handleA}
\end{table}


\paragraph{Note} The fact that elements of the set of all results $ \left\{ \mathsf{Metric}_p\!\left(\hat{y}_i(v)\right) : p\in P, v \in V,~0 \leq i < n \right\}$ may have to be excluded implies that the order in which results are averaged over phases, videos, and runs can make a difference.
Intuitively, if we average over a subset of the results first but have to exclude elements from this subset, then the remaining elements in the subset will be weighted a bit more in the overall result. 

This effect can be seen in the following example. Let us assume that we have phase-wise video-wise results for three phases $p_0, p_1, p_2$ and three videos $v_0, v_1, v_2$, summarized in matrix $X = \left( x_{p_j}(v_i) \right)_{\substack{0 \leq i < 3\\ 0 \leq j < 3}}$, where
\vspace{-1.5em}
$$
 X = \begin{pNiceMatrix}[first-row,first-col]
    & p_0 & p_1 & p_2\\
v_0 & 0.1   & 0.2   & 0.3  \\
v_1 & 0.1   & 0.2   & \text{n/a}  \\
v_2 & 0.1   & \text{n/a}   & 0.3  \\
\end{pNiceMatrix}.
$$
Computing $\mathrm{mean} \left\{ \mathrm{mean} \left\{ x_p(v) : p \in P \right\} : v \in V \right\}$, \emph{i.e.}, averaging over phases first, then over videos, yields:
$$\mathrm{mean} \left\{ \frac{0.1 + 0.2 + 0.3}{3}, \frac{0.1 + 0.2}{2}, \frac{0.1 + 0.3}{2} \right\} = \mathrm{mean} \left\{ 0.2, 0.15, 0.2 \right\} = \frac{0.2 + 0.15 + 0.2}{3} = 0.1833$$
Computing $\mathrm{mean} \left\{ \mathrm{mean} \left\{ x_p(v) : v \in V \right\} : p \in P \right\}$, \emph{i.e.}, averaging over videos first, then over phases, yields:
$$ \mathrm{mean} \left\{ \frac{0.1 + 0.1 + 0.1}{3}, \frac{0.2 + 0.2}{2}, \frac{0.3 + 0.3}{2} \right\} = \mathrm{mean} \left\{ 0.1, 0.2, 0.3 \right\} = \frac{0.1 + 0.2 + 0.3}{3} = 0.2$$
Finally, computing the average over all elements in the set, without any specific order, yields:
$$\mathrm{mean} \left\{ x_p(v) : p \in P, v \in V \right\} =   \frac{0.1 + 0.2 + 0.3 + 0.1 + 0.2 + 0.1 + 0.3}{7} = 0.1857$$

There can be more than one reasonable order of averaging results, often also depending on the use case. When reporting results for the macro-averaged metrics, for example, the numbers are averaged over phases first. 
In general, we propose to average over all elements at once, thus refraining from defining a specific order. This also means that all valid values in the set of results are weighted equally.

\subsubsection{Frame-wise evaluation metrics}
\label{sec:frame-wise}
Another strategy to avoid issues with undefined values is to compute the precision, recall, F1, and Jaccard scores over all video frames in the test set~$V$ instead of each video individually.\footnote{For example, \citet{rivoir2022pitfalls} calculate frame-wise F1 scores.}

To this end, the \emph{frame-wise} confusion matrix for the results obtained in the $i$-th experimental run is defined as
$$\EuScript{C}\!\left({\hat{y}_i(V)}\right) := \sum_{v \in V} C\!\left(\hat{y}_i(v)\right)\text{, where } 0 \leq i < n.$$
$\EuScript{C}\!\left({\hat{y}_i(V)}\right)$ counts the true positive, false positive, and false negative predictions for all video frames in the test set.

The phase-wise frame-wise evaluation metrics are computed as defined in section~\ref{sec:metrics}, but using the entries in the frame-wise confusion matrix $\EuScript{C}\!\left({\hat{y}_i(V)}\right)$:
\begin{align*}
    f\text{-}\mathsf{Precision}_p\!\left(\hat{y}_i(V)\right) &:= \frac{\EuScript{C}\!\left({\hat{y}_i(V)}\right)_{pp}}{\sum_q{\EuScript{C}\!\left({\hat{y}_i(V)}\right)_{qp}}} \\
    f\text{-}\mathsf{Recall}_p\!\left(\hat{y}_i(V)\right) &:= \frac{\EuScript{C}\!\left({\hat{y}_i(V)}\right)_{pp}}{\sum_q{\EuScript{C}\!\left({\hat{y}_i(V)}\right)_{pq}}}\\
    f\text{-}\mathsf{F1}_p\!\left(\hat{y}_i(V)\right) &:= \frac{2 \cdot \EuScript{C}\!\left({\hat{y}_i(V)}\right)_{pp}}{\sum_q{\EuScript{C}\!\left({\hat{y}_i(V)}\right)_{pq}} + \sum_q{\EuScript{C}\!\left({\hat{y}_i(V)}\right)_{qp}}} \\
    f\text{-}\mathsf{Jaccard}_p\!\left(\hat{y}_i(V)\right) &:= \frac{\EuScript{C}\!\left({\hat{y}_i(V)}\right)_{pp}}{\sum_q{\EuScript{C}\!\left({\hat{y}_i(V)}\right)_{pq}} + \sum_q{\EuScript{C}\!\left({\hat{y}_i(V)}\right)_{qp}} - \EuScript{C}\!\left({\hat{y}_i(V)}\right)_{pp}}
\end{align*}

These metrics, calculated for $n$ experimental runs, can be summarized as follows:
\begin{itemize}
    \item Overall mean: $M(f\text{-}\mathsf{Metric}) := \mathrm{mean} \left\{ f\text{-}\mathsf{Metric}_p\!\left(\hat{y}_i(V)\right) : p \in P,~0 \leq i < n \right\}$ 
    \item Variation over phases: $SD_{P}(f\text{-}\mathsf{Metric}) := \mathrm{std} \left\{ \mathrm{mean} \left\{ f\text{-}\mathsf{Metric}_p\!\left(\hat{y}_i(V)\right) : 0 \leq i < n \right\} : p \in P \right\}$
    \item Variation over runs: $SD_{R}(f\text{-}\mathsf{Metric}) := \mathrm{std} \left\{ \mathrm{mean} \left\{ f\text{-}\mathsf{Metric}_p\!\left(\hat{y}_i(V)\right) : p \in P \right\} : 0 \leq i < n \right\}$ 
    \item Phase-wise mean and variation over runs for each phase $p \in P$:
    \begin{itemize}
        \item $M(f\text{-}\mathsf{Metric}_p) := \mathrm{mean} \left\{ f\text{-}\mathsf{Metric}_p\!\left(\hat{y}_i(V)\right) : 0 \leq i < n \right\}$
        \item $SD_{R}(f\text{-}\mathsf{Metric}_p) := \mathrm{std} \left\{ f\text{-}\mathsf{Metric}_p\!\left(\hat{y}_i(V)\right) : 0 \leq i < n \right\}$
    \end{itemize}
\end{itemize}

\subsubsection{Relaxed evaluation metrics}
\label{subsec:relaxed}
 
When annotating surgical workflow, it may not always be absolutely clear at what time step one phase ends and the next phase begins. Therefore, it seems reasonable to accept minor deviations in the predicted timing of phase transitions, \emph{i.e.}, transitioning a few time steps earlier or later. To this end, \emph{relaxed} video-wise evaluation metrics were proposed as part of the \emph{Modeling and Monitoring of Computer Assisted Interventions~(M2CAI)} workflow challenge in 2016\footnote{\url{http://camma.u-strasbg.fr/m2cai2016/index.php/program-challenge/}}. \cite{jin2021temporal}\footnote{\url{https://github.com/YuemingJin/TMRNet/tree/main/code/eval/result/matlab-eval}} provide an adaptation of the original MATLAB script for the Cholec80 data.

The main idea is to treat erroneous predictions more generously when they occur within the first or last $\omega$ time steps of an annotated phase segment: Within the first $\omega$ time steps of true phase segment $q$, predictions of phase $\hat{q} \neq q$ should be accepted if $(\hat{q}, q) \in \EuScript{T}$, meaning that it is possible that $\hat{q}$ immediately precedes $q$ in the Cholec80 workflow and thus, this could be a late transition to the correct phase. Similarly, within the last $\omega$ time steps, 
predictions of phase $\hat{q} \neq q$ should be accepted if $(q, \hat{q}) \in \EuScript{T}$, meaning that $\hat{q}$ can immediately follow $q$ in the Cholec80 workflow and thus, this could be an early transition to the next phase. 
Here, $\EuScript{T}$ is the set of valid phase transitions, see Fig.~\ref{fig:phase-order}. 
The matrices $\EuScript{R}_{\mathrm{start}}$ and $\EuScript{R}_{\mathrm{end}}$ (Fig.~\ref{fig:mat-relaxed}) present in detail which predictions $\hat{q} \neq q$ are accepted in the implementation by \cite{jin2021temporal}.

\begin{figure}[htb]
\def\arrvline{\hfil\kern\arraycolsep\vline\kern-\arraycolsep\hfilneg}
    \begin{subfigure}[h]{0.45\textwidth}
        \centering
        \begin{tabular}{rllllllll}
         & & \multicolumn{7}{c}{$\hat{q}$} \\   
         &       & 0     & 1     & 2     & 3     & 4     & 5     & 6     \\
        \cmidrule(r){3-9}
        \multirow{7}{5pt}{$q$} & 0 \arrvline  & $\;\!\cdot$    & $0$    & $0$    & $0$    & $0$    & $0$    & $0$     \\  
            & 1 \arrvline   & $\mathbf{1}$    & $\;\!\cdot$    & $0$    & $0$    & $0$    & $0$    & $0$     \\ 
            & 2 \arrvline   & $0$    & $\mathbf{1}$    & $\;\!\cdot$    & $0$    & $0$    & $0$    & $0$     \\ 
            & 3 \arrvline   & $0$    & $0$    & $\mathbf{1}$    & $\;\!\cdot$    & $0$    & $0$    & $0$     \\ 
            & 4 \arrvline   & $0$    & $0$    & $0$    & $\mathbf{1}$    & $\;\!\cdot$    & $\!\mathbf{\times}$    & $0$     \\ 
            & 5 \arrvline   & $0$    & $0$    & $0$    & $\mathbf{1}$    & $\mathbf{1}$    & $\;\!\cdot$    & $\!\mathbf{\times}$     \\ 
            & 6 \arrvline   & $0$    & $0$    & $0$    & $0$    & $\mathbf{1}$    & $\mathbf{1}$    & $\;\!\cdot$     \\ 
	    \end{tabular}
       \caption{$\EuScript{R}_{\mathrm{start}}$} 
       \label{tab:firstw}
    \end{subfigure}
    \begin{subfigure}[h]{0.45\textwidth}
        \centering
        \begin{tabular}{rllllllll}
         & & \multicolumn{7}{c}{$\hat{q}$} \\   
         &       & 0     & 1     & 2     & 3     & 4     & 5     & 6     \\
        \cmidrule(r){3-9}
        \multirow{7}{5pt}{$q$} & 0 \arrvline  & $\;\!\cdot$    & $\mathbf{1}$    & $0$    & $0$    & $0$    & $0$    & $0$     \\  
            & 1 \arrvline   & 0    & $\;\!\cdot$    & $\mathbf{1}$    & $0$    & $0$    & $0$    & $0$     \\ 
            & 2 \arrvline   & $0$    & $0$    & $\;\!\cdot$    & $\mathbf{1}$    & $0$    & $0$    & $0$     \\ 
            & 3 \arrvline   & $0$    & $0$    & $0$    & $\;\!\cdot$    & $\mathbf{1}$    & $\mathbf{1}$    & $0$     \\ 
            & 4 \arrvline   & $0$    & $0$    & $0$    & $0$    & $\;\!\cdot$    & $\mathbf{1}$    & $\mathbf{1}$     \\ 
            & 5 \arrvline   & $0$    & $0$    & $0$    & $0$    & $\!\mathbf{\times}$    & $\;\!\cdot$    & $\mathbf{1}$     \\ 
            & 6 \arrvline   & $0$    & $0$    & $0$    & $0$    & $0$    & $\!\mathbf{\times}$    & $\;\!\cdot$     \\ 
	    \end{tabular}
        \caption{$\EuScript{R}_{\mathrm{end}}$} 
        \label{tab:lastw}
     \end{subfigure}
\caption{Matrices $\EuScript{R}_{\boldsymbol{\star}}$ to define which predictions $\hat{q} \neq q$ are accepted (a) within the first $\omega$ time steps and (b) within the last $\omega$ time steps of phase segment $q$. In case of acceptance, $\EuScript{R}_{\boldsymbol{\star}}[q, \hat{q}] = \mathbf{1}$, otherwise $\EuScript{R}_{\boldsymbol{\star}}[q,\hat{q}] = 0$. $\EuScript{R}_{\boldsymbol{\star}}[q,\hat{q}] = \times$ means that $\hat{q}$ should be accepted since (a) $(\hat{q}, q) \in \EuScript{T}$ or (b) $(q, \hat{q}) \in \EuScript{T}$, but is not considered in the script by \cite{jin2021temporal}.}
\label{fig:mat-relaxed}
\end{figure}

Formally, true positive predictions for phase $p \in P$ are counted under \emph{relaxed} conditions as 
$$\EuScript{R}\text{-}\mathsf{TP}_p\!\left(\hat{y}(v)\right) := \lvert \{t : \left(y(v)_t = p ~\mathrm{or}~ \hat{y}(v)_t = p\right) ~\mathrm{and}~ \EuScript{R}\!\left(\hat{y}(v)\right)_t = \mathrm{True}, 0 \leq t < T \} \rvert.$$
Here, $\EuScript{R}\!\left(\hat{y}(v)\right)_t$ specifies whether the prediction at time $t$ is considered correct under relaxed conditions\footnote{Here, $\mathrm{start}_q$ refers to the start time of phase segment $q$ and $\mathrm{end}_q$ refers to the end time of phase segment $q$.}:
\begin{align*}
\EuScript{R}\!\left(\hat{y}(v)\right)_t :\Leftrightarrow &\left(\hat{y}(v)_t = y(v)_t\right) \mathrm{~or} \\
 &\left(\exists q \in y(v): \mathrm{start}_q \leq t < (\mathrm{start}_q + \omega) ~\mathrm{and}~ \EuScript{R}_\mathrm{start}[q, \hat{y}(v)_t] = 1 \right) \mathrm{~or} \\
 &\left(\exists q \in y(v): (\mathrm{end}_q - \omega) < t \leq \mathrm{end}_q ~\mathrm{and}~ \EuScript{R}_\mathrm{end}[q, \hat{y}(v)_t] = 1 \right)\\
\end{align*}
Clearly, $\EuScript{R}\text{-}\mathsf{TP}_p \geq \mathsf{TP}_p$, and $\EuScript{R}\text{-}\mathsf{TP}_p$ reduces to $\mathsf{TP}_p$ when the relaxed conditions around phase transitions or boundaries are removed,
meaning that strictly $\EuScript{R}\!\left(\hat{y}(v)\right)_t \Leftrightarrow \hat{y}(v)_t = y(v)_t$.

The following relaxed metrics are defined:
\begin{align*}
    \EuScript{R}\text{-}\mathsf{Jaccard}_p\!\left(\hat{y}(v)\right) &:= \frac{\EuScript{R}\text{-}\mathsf{TP}_p\!\left(\hat{y}(v)\right)}{\lvert \{t : y(v)_t = p ~\mathrm{or}~ \hat{y}(v)_t = p, 0 \leq t < T\} \rvert} = \frac{\EuScript{R}\text{-}\mathsf{TP}_p}{\mathsf{TP}_p + \mathsf{FP}_p + \mathsf{FN}_p} \\
    \EuScript{R}\text{-}\mathsf{Precision}_p\!\left(\hat{y}(v)\right) &:= \frac{\EuScript{R}\text{-}\mathsf{TP}_p\!\left(\hat{y}(v)\right)}{\lvert \{t : \hat{y}(v)_t = p, 0 \leq t < T\} \rvert} = \frac{\EuScript{R}\text{-}\mathsf{TP}_p}{\mathsf{TP}_p + \mathsf{FP}_p} \\
    \EuScript{R}\text{-}\mathsf{Recall}_p\!\left(\hat{y}(v)\right) &:= \frac{\EuScript{R}\text{-}\mathsf{TP}_p\!\left(\hat{y}(v)\right)}{\lvert \{t : y(v)_t = p , 0 \leq t < T\} \rvert} = \frac{\EuScript{R}\text{-}\mathsf{TP}_p}{\mathsf{TP}_p + \mathsf{FN}_p} \\
    \EuScript{R}\text{-}\mathsf{Accuracy}\!\left(\hat{y}(v)\right) &:= \frac{\lvert \{t : \EuScript{R}\!\left(\hat{y}(v)\right)_t = \mathrm{True}, 0 \leq t < T \} \rvert}{T}
\end{align*}

\begin{figure}[t]
	\centering
	\begin{tabular}{l|cccccccccccccccccccc}
         & \multicolumn{20}{c}{$t \longrightarrow$} \\
        $y(v)_t$                            & $\cdots$   & 3 & \cellcolor{gray!20} 3 & \cellcolor{gray!20} 3 & \cellcolor{gray!40} 4 & \cellcolor{gray!40} 4 & 4 & 4 & \cellcolor{gray!20} 4 & \cellcolor{gray!20} 4 & \cellcolor{gray!40} 5 & \cellcolor{gray!40} 5 & 5 & 5 & \cellcolor{gray!20} 5 & \cellcolor{gray!20} 5 & \cellcolor{gray!40} 6 & \cellcolor{gray!40} 6 & 6 & $\cdots$ \\
        $\hat{y}(v)_t$                      & $\cdots$   & 3 & \cellcolor{gray!20} 5 & \cellcolor{gray!20} 4 & \cellcolor{gray!40} 4 & \cellcolor{gray!40} 3 & 3 & 3 & \cellcolor{gray!20} 4 & \cellcolor{gray!20} 6 & \cellcolor{gray!40} 3 & \cellcolor{gray!40} 4 & 4 & 6 & \cellcolor{gray!20} 5 & \cellcolor{gray!20} 6 & \cellcolor{gray!40} 5 & \cellcolor{gray!40} 4 & 6 & $\cdots$ \\
        $\EuScript{R}\!\left(\hat{y}(v)\right)_t$ & $\cdots$ & $\mathrm{T}$ & $\mathrm{T}$ & $\mathrm{T}$ & $\mathrm{T}$ & $\mathrm{T}$ & $\mathrm{F}$ & $\mathrm{F}$ & $\mathrm{T}$ & $\mathrm{T}$ & $\mathrm{T}$ & $\mathrm{T}$ & $\mathrm{F}$ & $\mathrm{F}$ & $\mathrm{T}$ & $\mathrm{T}$ & $\mathrm{T}$ & $\mathrm{T}$ & $\mathrm{T}$ & $\cdots$ \\

	\end{tabular}
 	\caption{Part of an exemplary annotation $y(v)$ and prediction $\hat{y}(v)$ to illustrate cases in which wrong predictions are considered correct, \emph{i.e.}, $\EuScript{R}\!\left(\hat{y}(v)\right)_t = \mathrm{True}$. In this example, $\omega$ is set to 2. We highlight the \colorbox{gray!40}{first} and \colorbox{gray!20}{last} $\omega$~time steps of each annotated phase segment. $\mathrm{True}$ is abbreviated as $\mathrm{T}$ and $\mathrm{False}$ as $\mathrm{F}$. }
    \label{fig:relaxed}
\end{figure}

\paragraph{Properties of relaxed evaluation metrics}
Relaxed evaluation metrics tolerate relatively noisy predictions. An example is depicted in Fig.~\ref{fig:relaxed}. Here, we can observe that the prediction of phase~4 at the end of phase~3 is excused as an early transition from phase~3 to phase~4. A few time steps later, the prediction of phase~3 at the beginning of annotated phase~4 is also accepted, now assuming a late transition from phase~3 to phase~4, which is somewhat contradictory. Also, the known order of the annotated phase segments is not taken into account. For example, it is clear that phase~4 is followed immediately by phase~5 in the example. Still, a prediction of phase~6 at the end of phase~4 and phase~3 at the beginning of phase~5 is accepted as well.  

For further demonstration, we calculate the relaxed metrics for phase~4 in the example (Fig.~\ref{fig:relaxed}), assuming that there are no further predictions of phase~4 in the parts of $\hat{y}(v)$ that are not depicted. By counting, we find that: \\
$\EuScript{R}\text{-}\mathsf{TP}_{p=4}\!\left(\hat{y}(v)\right) = 7,~ \lvert \{t : y(v)_t = 4 ~\mathrm{or}~ \hat{y}(v)_t = 4\} \rvert = 10,~ \lvert \{t : \hat{y}(v)_t = 4\} \rvert = 6,~ \lvert \{t : y(v)_t = 4\} \rvert = 6$. \\
$\text{It follows that}~ \EuScript{R}\text{-}\mathsf{Jaccard}_{p=4}\!\left(\hat{y}(v)\right) = 0.7,~ \EuScript{R}\text{-}\mathsf{Precision}_{p=4}\!\left(\hat{y}(v)\right) = 1.167,~ \EuScript{R}\text{-}\mathsf{Recall}_{p=4}\!\left(\hat{y}(v)\right) = 1.167$.

Obviously, the chosen example is extreme since the larger parts of the annotated phase segments 4 and 5 are handled in a relaxed manner. In the existing implementations, $\omega$ is set to $10\,s$, meaning that $20\,s$ of each annotated phase segment are evaluated under relaxed conditions. For phases 1 and 3, the relaxed phase-wise metrics will not differ much from the standard metrics because the annotated phase segments are much longer than $20\,s$. 
However, the underrepresented phases in Cholec80 can be quite short (median duration $< 140\,s$, see Fig.~\ref{fig:phase-duration}), meaning that calculating metrics with relaxed boundaries can make more of a difference. Consequently, relaxed evaluation metrics weight errors on short phases less while standard $\mathsf{Jaccard}$, $\mathsf{Precision}$, and $\mathsf{Recall}$ weight all phases equally. 
Still, the underrepresented phases are more difficult to predict, first because there are fewer example video frames available for training and second because the progress of surgical phases after phase 3 is not necessarily linear, see Fig.~\ref{fig:phase-order}.

\paragraph{Issue with overly high precision and recall scores}
As calculated above, the relaxed precision and recall scores can actually exceed $1$. The reason for this is that the relaxed true positives $\EuScript{R}\text{-}\mathsf{TP}_p\!\left(\hat{y}(v)\right)$ are counted for all time steps at which phase $p$ is either annotated or predicted. Yet, in the case of precision, it would be more reasonable to consider only the time steps at which phase $p$ is predicted and then count how many predictions at these time steps are considered correct under relaxed conditions. Analogously, in the case of recall, only the time steps at which phase $p$ is annotated should be considered when counting true positives. 
In contrast, the implementation provided by \cite{jin2021temporal} 
truncates all scores to a maximum value of 1 (or $100\,\%$) before computing the summary metrics.

\begin{lstlisting}[caption={Code section from \href{https://github.com/YuemingJin/TMRNet/blob/main/code/eval/result/matlab-eval/Evaluate.m}{Evaluate.m}. Here, \texttt{t} refers to~$\omega$. Note that the MATLAB script uses 1-based indexing, also when referring to surgical phases.},label=code]
% relaxed boundary
% revised for cholec80 dataset !!!!!!!!!!!
if(iPhase == 4 || iPhase == 5) % Gallbladder dissection and packaging might jump between two phases
    curDiff(curDiff(1:t)==-1) = 0; % late transition
    curDiff(curDiff(end-t+1:end)==1 | curDiff(end-t+1:end)==2) = 0; % early transition % 5 can be predicted as 6/7 at the end > 5 followed by 6/7
elseif(iPhase == 6 || iPhase == 7) % Gallbladder dissection might jump between two phases
    curDiff(curDiff(1:t)==-1 | curDiff(1:t)==-2) = 0; % late transition
    curDiff(curDiff(end-t+1:end)==1 | curDiff(end-t+1:end)==2) = 0; % early transition
else
    % general situation
    curDiff(curDiff(1:t)==-1) = 0; % late transition
    curDiff(curDiff(end-t+1:end)==1) = 0; % early transition
end
\end{lstlisting}

\paragraph{Issue with existing MATLAB scripts}
Notably, there is a bug in the implementation\footnote{The problem can be found both in the MATLAB scripts that were disseminated with the M2CAI challenge and in the adjusted scripts provided by \cite{jin2021temporal}.} of the relaxed metrics that alters the computation of $\EuScript{R}\!\left(\hat{y}(v)\right)_t$. The problem can be seen in Listing~\ref{code}.

\begin{figure}[htb]
     \centering
     \begin{subfigure}[t]{0.495\textwidth}         
        \centering
        \captionsetup{margin=1.cm, indention=-1.cm}
        \begin{adjustbox}{clip,trim=2cm 3.5cm 9cm 2.5cm,width={0.99\textwidth},keepaspectratio}
        \includegraphics{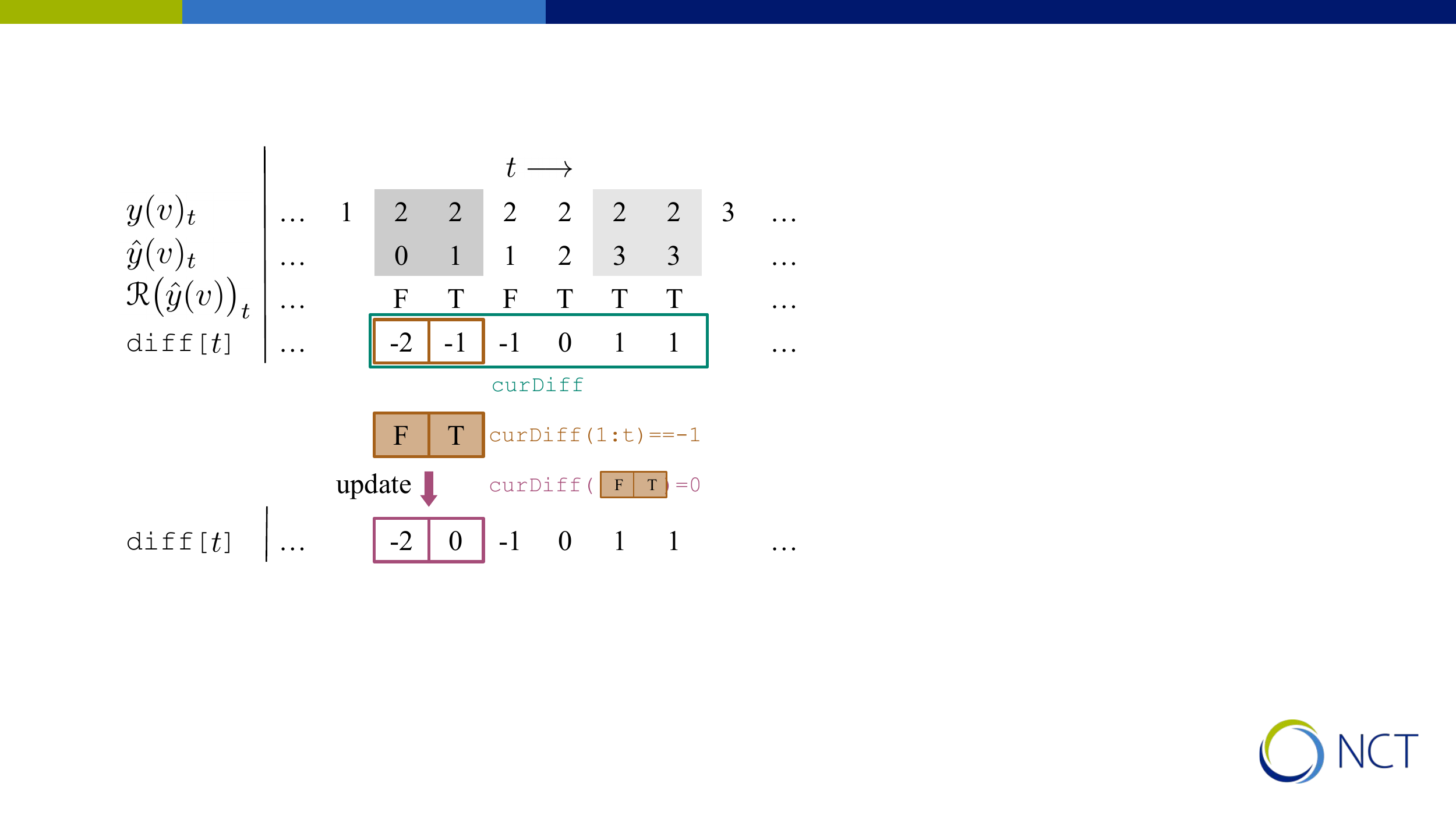}
        \end{adjustbox}
        \cprotect\caption{Update implemented in line 11. If the difference between $\hat{y}(v)_t$ and $y(v)_t$ equals -1, then this means that the phase that immediately \emph{precedes} the annotated phase in the Cholec80 workflow is predicted. This is tolerated for the \emph{first }$\omega$ time steps in the annotated phase segment, meaning that the corresponding \verb|diff[|$t$\verb|]| is set to zero. }
        \label{fig:relaxed_bug_first}
     \end{subfigure}
     \hfill
     \begin{subfigure}[t]{0.495\textwidth}
        \centering
        \captionsetup{indention=-0.75cm}
        \begin{adjustbox}{clip,trim=2cm 3.5cm 9cm 2.5cm,width={0.99\textwidth},keepaspectratio}
        \includegraphics{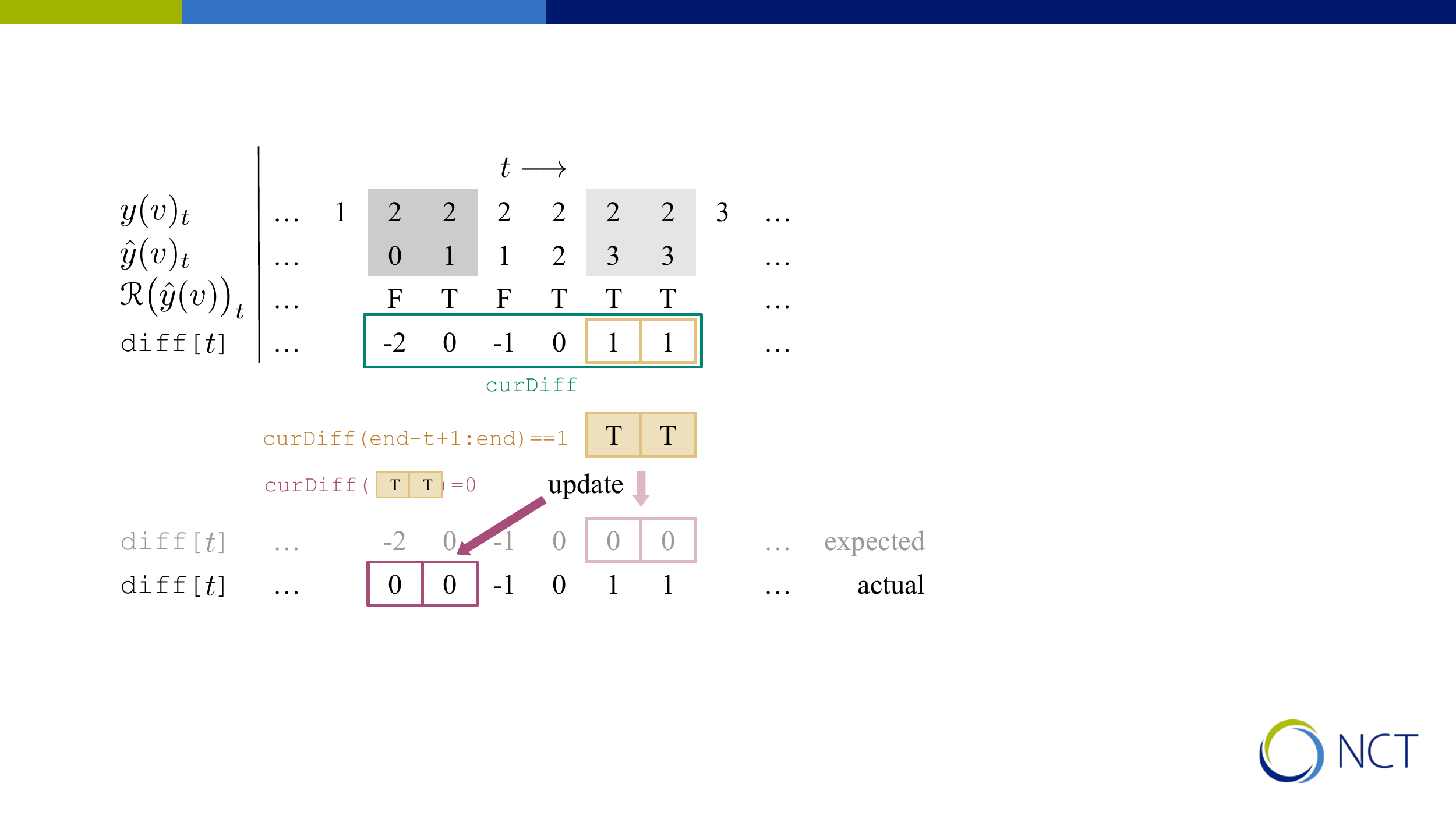}
        \end{adjustbox}
        \cprotect\caption{Update implemented in line 12. If the difference between $\hat{y}(v)_t$ and $y(v)_t$ equals 1, then this means that the phase that immediately \emph{follows} the annotated phase in the Cholec80 workflow is predicted. This should be tolerated for the \emph{last} $\omega$ time steps in the annotated phase segment. 
        Due to MATLAB semantics, however, the computed \colorbox{tabcolor1!50}{Boolean array of length $\omega$} will be used to select (and reset) array positions within the \textbf{first} $\omega$ positions in \verb|curDiff|.}
        \label{fig:relaxed_bug_last}
     \end{subfigure}
     \hfill
     \cprotect\caption{Illustration of how the \verb|diff| array is updated in the MATLAB script when \verb|curDiff| refers to the segment that is annotated as phase~2 (\emph{i.e.} \verb|iPhase == 3|). 
     In this example, $\omega$ (\emph{i.e.} \verb|t|) is set to 2. We highlight the \colorbox{gray!40}{first} and \colorbox{gray!20}{last} $\omega$~time steps of the phase segment. Note that $t$, not to be confused with \verb|t|, refers to the time step as usual.
     After all updates, the predictions at the time steps $t$ for which \verb|diff[|$t$\verb|]|$= 0$ will be considered correct under relaxed conditions.
     }     
     \label{fig:bug_relaxed}
\end{figure}

To understand the code, it is necessary to know that there is an array \verb|diff| defined, which is initialized as $\hat{y}(v) - y(v)$, see Fig.~\ref{fig:bug_relaxed} for an example. In the first part of the script, \verb|diff| is manipulated such that it encodes $\EuScript{R}\!\left(\hat{y}(v)\right)_t$, namely, $\EuScript{R}\!\left(\hat{y}(v)\right)_t \Leftrightarrow \verb|diff[|t\verb|]| = 0$. 
For adjusting \verb|diff|, the script iterates through all phases \verb|iPhase|, looking at the part of \verb|diff| that corresponds to the segment that is annotated as \verb|iPhase|, namely, \verb|curDiff|$:=$\verb|diff[|$\mathrm{start}_{\verb|iPhase|} \cdots \mathrm{end}_{\verb|iPhase|}$\verb|]|.
Then, entries in \verb|curDiff| (and thus \verb|diff|, accordingly) are set to zero if certain conditions are met, meaning that some erroneous predictions are ignored.

To see how the first three phases are handled, see lines 10-12. If for $0 \leq j < \omega$ it holds that \verb|curDiff[|$j$\verb|]|~$ = -1$, then  \verb|curDiff[|$j$\verb|]| is set to zero. Thus, if within the first $\omega$ frames of phase 1 or 2 the predicted phase is 0 or 1, respectively, then this error is ignored (see Fig.~\ref{fig:relaxed_bug_first} for an illustration).  This is expected behavior. The bug is in line 12 (lines 5 and 8 analogously), see Fig.~\ref{fig:relaxed_bug_last}.
Expectedly, line~12 would have the effect that if $0 \leq j < \omega$ and \verb|curDiff[|$\verb|len(curDiff)| - \omega + j$\verb|]|~$= 1$, then \verb|curDiff[|$\verb|len(curDiff)| - \omega + j$\verb|]| is set to zero. This would imply that if within the last $\omega$ frames of phase 0, 1, or 2 the predicted phase is 1, 2, or 3, respectively, then this error would be tolerated.  
However, the MATLAB expression \verb|curDiff(end-t+1:end)==1| returns a Boolean array of length $\omega$, which is then used to manipulate the \textbf{first} $\omega$ positions in \verb|curDiff|.
Consequently, \verb|curDiff[|$j$\verb|]| will be set to zero if \verb|curDiff[|$\verb|len(curDiff)| - \omega + j$\verb|]|~$= 1$. To clarify: If a reasonable error within the last $\omega$ frames of a phase occurs, then this error will never be ignored. Instead, any error, reasonable or not, at the corresponding position within the first $\omega$ frames will be accepted.
This behavior is clearly unintended and can most likely be attributed to the use of ambiguous notation in MATLAB. 

\section{Worked example}
\label{sec:example}

To showcase the effect of sometimes very subtle differences in model evaluation, we trained a baseline model\footnote{
The baseline model is a ResNet-LSTM that was trained end to end on short video sequences of 8~frames, where video frames were extracted at a temporal resolution of 1~fps. 
For training, we used the regular cross-entropy loss computed on the predictions for all eight frames in the sequence. At test time, we obtained the prediction at time~$t$ by applying the ResNet-LSTM to the sequence $\left(v_{t-7}, ..., v_{t-1}, v_t\right)$ and taking the final prediction.

The ResNet-50 was initialized with the weights obtained by pre-training on ImageNet \citep{deng2009imagenet} while orthogonal initialization \citep{saxe2013exact} was applied to the LSTM. Further, the weights of the three initial ResNet layers (conv1, conv2, and conv3) were frozen. 
The model was trained with the AdamW~optimizer \citep{loshchilov2017decoupled} for 1600~epochs, using a one cycle learning rate schedule \citep{leslie2019super} with a maximum learning rate of $3\times10^{-4}$ and a constant weight decay of $0.01$. 
In each training epoch, five sequences from each phase and one sequence around each phase transition were sampled from each training video, shuffled, and then processed in batches of 80~sequences. 
For data augmentation, we used the Albumentations library \citep{info11020125} to apply a number of randomized transformations  to each video sequence, including color and spatial transformations, blurring, and horizontal flipping.
Basically, our baseline is a re-implementation of SV-RCNet \citep{jin2017sv}, using an adjusted training strategy.  
} on the Cholec80 data and computed the evaluation metrics on the test videos. The evaluation code is publicly available at \url{https://gitlab.com/nct_tso_public/phasemetrics}.

During model development, we trained on 32 videos and validated on 8 videos to tune hyperparameters such as learning rate and the number of training steps. Then, with a fixed set of hyperparameters, we repeated the model training with different random seeds to collect the results of five experimental runs on the test set of 40 videos. The summarized results are reported in Table~\ref{tab:results_32-8-40_1} and Table~\ref{tab:results_32-8-40_2}.

\definecolor{lightgray}{gray}{0.95}
\definecolor{mygray}{gray}{0.5}
\newcolumntype{g}{>{\color{mygray}}l}
\begin{table}[htb]
\caption{Results on the 32:8:40 data split. For each metric, we report the mean~$M$, standard deviation over videos~$SD_V$, standard deviation over phases~$SD_P$, and standard deviation over runs~$SD_R$. Results are computed using strategies~$\mathsf{A}$ and $\mathsf{B}$ for handling undefined values.}
\label{tab:results_32-8-40_1}
\centerline{
\resizebox{1.0\textwidth}{!}{%
\begin{tabular}{clggglggglggglggg}
    \toprule
    & \multicolumn{4}{l}{$\mathsf{Precision}$} & \multicolumn{4}{l}{$\mathsf{Recall}$} & \multicolumn{4}{l}{$\mathsf{Jaccard}$} & \multicolumn{4}{l}{$\mathsf{F1}$}  \\
          & $M$             & $SD_V$    & $SD_P$    & $SD_R$     &$M$             & $SD_V$  & $SD_P$         & $SD_R$    & $M$             & $SD_V$  & $SD_P$      & $SD_R$     & $M$             & $SD_V$        & $SD_P$      & $SD_R$         \\
    \midrule
$\mathsf{A}$ & .826      & .075   & .082  & .005     & .805     & .066   & .107     & .006    & .681     & .093  & .117   & .006     & .785    & .079  & .102  & .005   \\
$\mathsf{B}$ & .839      & .070    & .057  & .006     & .805     & .066    & .107  & .006    & .692     & .092    & .103  & .007     & .797    & .074   & .084 & .006    \\      
    \bottomrule
\end{tabular}
}
}
\end{table}

\begin{table}[htb]
\caption{Results on the 32:8:40 data split. For each metric, we report the mean~$M$, standard deviation over videos~$SD_V$, and standard deviation over runs~$SD_R$. If applicable, results are computed using strategies~$\mathsf{A}$ and $\mathsf{B}$ for handling undefined values.}
\label{tab:results_32-8-40_2}
\centering
\begin{tabular}{clgglgglgg}
    \toprule
    & \multicolumn{3}{l}{$\mathsf{Macro\,F1}$} & \multicolumn{3}{l}{$\mathbf{Macro\,F1}$} & \multicolumn{3}{l}{$\mathsf{Accuracy}$} \\
           & $M$             & $SD_V$         & $SD_R$   & $M$             & $SD_V$         & $SD_R$  & $M$             & $SD_V$         & $SD_R$         \\
    \midrule
$\mathsf{A}$  & .785    & .079   & .005   & .814       & .063      & .005   & .865      & .066      & .003       \\
$\mathsf{B}$  & .797    & .074   & .005   & .821       & .062      & .005    &        &      &  \\      
    \bottomrule
\end{tabular}
\end{table}

For each metric, we compute the overall mean, standard deviation over videos, standard deviation over five experimental runs and, if applicable, standard deviation over phases.
We can see that the variation over runs is relatively small compared to the variation over videos or phases. Also, the standard deviation over phases can exceed 0.1 for some metrics, meaning that there are considerable differences in how well the model performs on different phases.

The strategy for handling undefined values -- or, more generally, the cases where a phase is missing in the video annotation -- can make a notable difference when computing and summarizing phase-wise video-wise evaluation metrics. 
More specifically, excluding all phase-wise results when the phase is missing in the video annotation ($\mathsf{Strategy\,B}$) leads to larger numbers compared to excluding undefined values only ($\mathsf{Strategy\,A}$). Only $\mathsf{Recall}$ is robust to the choice of either $\mathsf{A}$ or $\mathsf{B}$ because it is \emph{undefined} \emph{iff} the phase is missing in the video annotation. Interestingly, macro-averaged recall has also been proposed for measuring \emph{Balanced Accuracy} \citep{guyon2015design}.
Besides that, we can also see that computing $\mathbf{Macro\,F1}$ instead of $\mathsf{Macro\,F1}$  leads to considerably larger numbers. As expected, $\mathit{F1}$, \emph{i.e.}, the harmonic mean of mean precision and mean recall, is even a bit higher: 0.815 for $\mathsf{Strategy\,A}$ and 0.822 for $\mathsf{Strategy\,B}$.

\begin{figure}[t]
	\centering
    \begin{adjustbox}{clip,trim=0cm 0.35cm 0cm 1cm,width={0.9\textwidth},keepaspectratio}
    \input{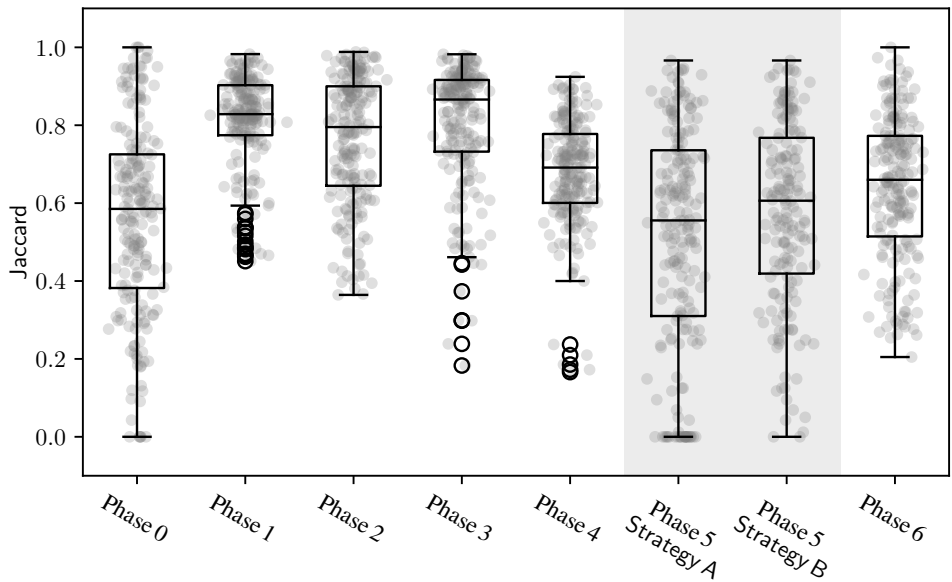}
    \end{adjustbox}
	\caption{Phase-wise $\mathsf{Jaccard}$ scores for all 40 test videos and all five experimental runs. Gray dots visualize individual data points, \emph{i.e.}, the phase-wise $\mathsf{Jaccard}$ score for one specific video in one run.}
	\label{fig:phasewise-jaccard}
\end{figure}

For insights into the model's performance on each individual surgical phase, we report the phase-wise evaluation metrics in Table~\ref{tab:32-8-40_phasewise}. In addition, we visualize the distributions of the phase-wise $\mathsf{Jaccard}$ scores on the test data, see Fig.~\ref{fig:phasewise-jaccard}.
In general, we can observe that phases 1 (\say{Calot triangle dissection}) and 3 (\say{Gallbladder dissection}), which are also the dominating phases in Cholec80, can be recognized relatively well in most videos.
In contrast, phases 0 (\say{Preparation}) and 5 (\say{Cleaning and coagulation}) cannot be recognized properly in many videos.
Besides that, we can see that the means over phase-wise means reported in Table~\ref{tab:32-8-40_phasewise} deviate slightly from the overall means reported in Table~\ref{tab:results_32-8-40_1}. The reason is that the order of averaging makes a difference when there are undefined values that need to be excluded (section~\ref{sec:undefined}). 

\begin{table}[t]
\caption{Phase-wise results on the 32:8:40 data split. For each metric, we report the mean~$M$, standard deviation over videos~$SD_V$, and standard deviation over runs~$SD_R$. In addition, we report $M_P$, which is the mean over all phase-wise means. 
Results that do not differ when using either $\mathsf{Strategy\,A}$ or $\mathsf{Strategy\,B}$ for handling undefined values are  reported only once for both strategies.}
\label{tab:32-8-40_phasewise}
\centering
\begin{tabular}{cclgglgglgglgg}
    \toprule
      && \multicolumn{3}{l}{$\mathsf{Precision}$} & \multicolumn{3}{l}{$\mathsf{Recall}$} & \multicolumn{3}{l}{$\mathsf{Jaccard}$} & \multicolumn{3}{l}{$\mathsf{F1}$} \\
Phase && $M$       & $SD_V$    & $SD_R$    & $M$       & $SD_V$    & $SD_R$   & $M$       & $SD_V$    & $SD_R$   & $M$       & $SD_V$    & $SD_R$ \\
    \midrule
0     && .856    & .202     & .025     & .623   & .253    & .021    & .561    & .236    & .019    & .683 & .219   & .021  \\
1     && .851    & .117     & .007     & .943   & .072    & .007    & .809    & .122    & .004    & .889 & .083   & .002  \\
2     && .906    & .099     & .013     & .836   & .165    & .017    & .762    & .158    & .006    & .854 & .111   & .005  \\
3     && .903    & .142     & .008     & .884   & .119    & .015    & .802    & .156    & .008    & .880 & .113   & .005  \\
4     && .808    & .148     & .015     & .821   & .102    & .011    & .679    & .132    & .005    & .800 & .108   & .005  \\
\rowcolor{lightgray} \cellcolor{lightgray} & $\mathsf{A}$  &  .669    & .339    & .018     &    &     &     & .510    & .291   & .008    & .619 & .309   & .011  \\
\rowcolor{lightgray} \multirow{-2}{*}{\cellcolor{lightgray}5}  &  $\mathsf{B}$   & .753    & .243     & .015    & \multirow{-2}{*}{.710}   & \multirow{-2}{*}{.240}    & \multirow{-2}{*}{.014}    & .574    & .234    & .009    & .697  & .217   & .010  \\
6     && .789    & .203     & .028     & .807   & .158    & .024    & .642    & .174    & .009    & .765 & .142   & .006 \\
\rowcolor{lightgray} \cellcolor{lightgray} & $\mathsf{A}$ & .826 & & & & & & .681 & & & .784 & & \\
\rowcolor{lightgray} \multirow{-2}{*}{\cellcolor{lightgray}$M_P$}  & $\mathsf{B}$ & .838 & & & \multirow{-2}{*}{.804} & & & .690 & & & .795 & & \\
    \bottomrule
\end{tabular}
\end{table}

Since phase~5 is the only phase that is missing in some of the test videos, only the phase-wise results for phase~5 
are affected by the choice of either $\mathsf{Strategy\,A}$ or $\mathsf{B}$ for handling undefined values.
More specifically, the scores for videos where phase~5 is missing are ignored when computing $\mathsf{Jaccard}_{p=5}$ with $\mathsf{Strategy\,B}$. 
When computing $\mathsf{Jaccard}_{p=5}$ with $\mathsf{Strategy\,A}$, however, the scores for videos  where phase~5 is missing are zero \emph{iff} there is at least one false positive prediction of phase~5 -- or \emph{undefined}, and thus ignored, otherwise.
This effect can also be seen in Fig.~\ref{fig:phasewise-jaccard}, where the distribution of scores for phase~5 and $\mathsf{Strategy\,A}$ simply includes a number of additional zeros in comparison to $\mathsf{Strategy\,B}$.\footnote{The same effect is to be expected for $\mathsf{Precision}$ and $\mathsf{F1}$. }
It seems reasonable to exclude these zeros as outliers.

For further analysis, we report the frame-wise results in Table~\ref{tab:32-8-40_framewise}. 
Obviously, the  frame-wise results are not directly comparable to the  video-wise results reported in Table~\ref{tab:32-8-40_phasewise}.
With video-wise calculation, the phase-wise metric is computed for each video and then averaged over all videos, weighting each video-wise result equally. This approach emphasizes that each phase should be recognized equally well in every video.  
In contrast, with frame-wise calculation, the phase-wise metric is computed once given the predictions on all frames in the test set. Therefore, videos and phase segments that contribute more frames, \emph{i.e.}, have a longer temporal duration, have a higher weight in the overall result.
Still, the ranking of individual phases is similar: Phase recognition works best for phases 1 and 3, and 
results are also promising for phases 2 (\say{Clipping and cutting}) and 4 (\say{Gallbladder packaging}). 
The results are worst for phases 0 and 6 (\say{Gallbladder retraction}).
Interestingly, the frame-wise results for phase~5 are better than for phases 0 and 6. One reason could be that it was especially difficult to recognize phase~5 in some specific videos, which weigh less in the frame-wise computation. 
To show which surgical phases are commonly confused with each other, we visualize the overall confusion matrix, 
computed over all video frames and experimental runs, in Fig.~\ref{fig:confusion-matrix}. 

\begin{table}[htb]
\caption{Frame-wise results on the 32:8:40 data split. For each metric, we report the overall and phase-wise mean~$M$ and standard deviation over runs~$SD_R$. In addition, we report the standard deviation over phases~$SD_P$.}
\label{tab:32-8-40_framewise}
\centering
\begin{tabular}{clglglglg}
    \toprule
      & \multicolumn{2}{l}{$f\text{-}\mathsf{Precision}$} & \multicolumn{2}{l}{$f\text{-}\mathsf{Recall}$} & \multicolumn{2}{l}{$f\text{-}\mathsf{Jaccard}$} & \multicolumn{2}{l}{$f\text{-}\mathsf{F1}$} \\
Phase & $M$         & $SD_R$    & $M$         & $SD_R$   & $M$           & $SD_R$   & $M$      & $SD_R$ \\
    \midrule
0     & .858     & .020    & .590     & .019    & .537    & .021   & .699  & .018  \\
1     & .853     & .010    & .927     & .011    & .800    & .005   & .889  & .003  \\
2     & .884     & .015    & .747     & .022    & .680    & .011   & .809  & .008  \\
3     & .900     & .010    & .878     & .013    & .800    & .006   & .889  & .004  \\
4     & .776     & .017    & .822     & .007    & .664    & .011   & .798  & .008  \\
5     & .773     & .008    & .743     & .013    & .610    & .008   & .757  & .006  \\
6     & .734     & .022    & .763     & .023    & .597    & .008   & .748  & .006 \\
\midrule 
 all   & .825     & .006    & .781     & .006    & .670    & .007   & .798  & .006 \\
\midrule
\color{mygray}$SD_P$ & \multicolumn{2}{c}{\color{mygray}.064} & \multicolumn{2}{c}{\color{mygray}.110} & \multicolumn{2}{c}{\color{mygray}.100} & \multicolumn{2}{c}{\color{mygray}.071} \\
    \bottomrule
\end{tabular}
\end{table}

\begin{figure}[htb]
     \centering
     \begin{subfigure}[b]{0.48\textwidth}         
        \centering
        \begin{adjustbox}{clip,trim=3.5cm 1cm 4cm 0cm,width={0.9\textwidth},keepaspectratio}
            \input{figures/confusion_matrix_norm-column.pgf}
        \end{adjustbox}
        \caption{Confusion matrix after column-wise normalization. The numbers on the diagonal show the $f\text{-}\mathsf{Precision}$ for each phase.}
     \end{subfigure}
     \hfill
     \begin{subfigure}[b]{0.48\textwidth}
        \centering
        \begin{adjustbox}{clip,trim=3.5cm 1cm 4cm 0cm,width={0.9\textwidth},keepaspectratio}
            \input{figures/confusion_matrix_norm-row.pgf}
        \end{adjustbox}
        \caption{Confusion matrix after row-wise normalization. The numbers on the diagonal show the $f\text{-}\mathsf{Recall}$ for each phase.}
     \end{subfigure}
     \hfill
     \caption{Overall confusion matrix $\underset{0 \leq i < 5}{\sum}~\underset{v \in V}{\sum}C\!\left(\hat{y}_i(v)\right)$.}
     \label{fig:confusion-matrix}
\end{figure}

Next, we computed the relaxed video-wise evaluation metrics (section~\ref{subsec:relaxed}) based on the original MATLAB scripts. Thus, we included the bug and the steps to trim overly high precision and recall scores.  As in the original script, $\mathsf{Strategy\,B}$ is used for handling undefined values. The results are presented in Table~\ref{tab:32-8-40_relaxed}. As expected, larger numbers are obtained when the evaluation metrics are computed with relaxed boundaries. 

For further insights, we report the relaxed phase-wise metrics in Table~\ref{tab:32-8-40_relaxed_phasewise}. Here, we additionally state the absolute difference~$\Delta$ between a relaxed result and the corresponding regular result, as reported in Table~\ref{tab:32-8-40_phasewise}. 
The phase-wise means $M_P$ are improved by an absolute $\Delta$ of at least~0.035 by computing results in a relaxed manner.
As anticipated, the differences are especially large on shorter phases. An exception is phase~0 because it is the very first phase in the Cholec80 workflow and thus, errors at the beginning of phase~0 will not be excused.

\begin{table}[t]
\caption{Relaxed results on the 32:8:40 data split. To be consistent with prior work, we report the mean $M_P$ and standard deviation over phases~$SD_P$ for $\mathsf{Precision}$, $\mathsf{Recall}$, and $\mathsf{Jaccard}$. $M_P$ is computed by averaging over videos and runs first and then over phases. For $\mathsf{Accuracy}$, we report the mean~$M$ and standard deviation over videos~$SD_V$. }
\label{tab:32-8-40_relaxed}
\centerline{
\begin{tabular}{lglglglg}
    \toprule
\multicolumn{2}{l}{$\EuScript{R}\text{-}\mathsf{Precision}$} & \multicolumn{2}{l}{$\EuScript{R}\text{-}\mathsf{Recall}$} & \multicolumn{2}{l}{$\EuScript{R}\text{-}\mathsf{Jaccard}$}  & \multicolumn{2}{l}{$\EuScript{R}\text{-}\mathsf{Accuracy}$}  \\
 $M_P$      &       $SD_P$   &$M_P$          & $SD_P$     & $M_P$             & $SD_P$      & $M$             & $SD_V$             \\
    \midrule
.873        & .043        & .840       & .100        & .726       &   .093  & .874  & .068        \\  
    \bottomrule
\end{tabular}
}
\end{table}

\begin{table}[t]
\caption{Relaxed phase-wise results on the 32:8:40 data split. For each metric, we report the mean~$M$, standard deviation over videos~$SD_V$, and standard deviation over runs~$SD_R$. In addition, we report $M_P$, which is the mean over all phase-wise means. $\Delta$ quantifies the absolute difference between a relaxed result and the corresponding \emph{regular} result. Here, for each metric, we set values in \textbf{bold} if they are greater than the $\Delta$ computed for the mean $M_P$.}
\label{tab:32-8-40_relaxed_phasewise}
\centering
\begin{tabular}{cllggllggllgg}
    \toprule
      & \multicolumn{4}{l}{$\EuScript{R}\text{-}\mathsf{Precision}$} & \multicolumn{4}{l}{$\EuScript{R}\text{-}\mathsf{Recall}$} & \multicolumn{4}{l}{$\mathsf{\EuScript{R}\text{-}Jaccard}$} \\
Phase & $M$ & $\Delta$      & $SD_V$    & $SD_R$    & $M$  & $\Delta$     & $SD_V$    & $SD_R$   & $M$ & $\Delta$      & $SD_V$    & $SD_R$   \\
    \midrule
0     & .891  & .035   & .149     & .017     & .649 & .026  & .228    & .018    & .588  & .027   & .216    & .020      \\
1     & .857 & .006   & .117     & .007     & .949  & .006  & .072    & .007    & .815  & .006  & .123    & .004      \\
2     & .939 & .033  & .102     & .012     & .870   & .034  & .167    & .014    & .810  & \textbf{.048}  & .176    & .004     \\
3     & .914 & .011   & .137     & .008     & .897  & .013  & .122    & .014    & .813  & .011  & .154    & .007     \\
4     & .857 & \textbf{.049}   & .140     & .012     & .877   & \textbf{.056}  & .095    & .011    & .728 & \textbf{.049}   & .127    & .005    \\
5     & .827 & \textbf{.074}   & .215     & .007     & .768   & \textbf{.058}  & .215    & .016    & .627 & \textbf{.053}   & .223    & .010      \\
6     & .828 & \textbf{.039}   & .198     & .025     & .873   & \textbf{.066} & .150    & .015    & .700  & \textbf{.058}  & .184    & .013     \\
$M_P$ & .873 & .035 & & & .840 & .036 & & & .726 & .036 & & \\
    \bottomrule
\end{tabular}
\end{table}

Next, we report the results for the best and worst out of all five experimental runs in Table~\ref{tab:32-8-40_best-worst}. Clearly, there can be notable differences between experimental runs, which, however, need to be attributed to pure randomness or chance. For this reason, it is recommended to summarize the results over several experimental runs, as done in Tables~\ref{tab:results_32-8-40_1} and~\ref{tab:results_32-8-40_2}.

\begin{table}[htb]
\caption{Results on the 32:8:40 data split, computed for the best and the worst experimental run. For each metric, we report the mean~$M$, standard deviation over videos~$SD_V$, and, if applicable, standard deviation over phases~$SD_P$.  $\mathsf{Strategy~B}$ is used for handling undefined values.}
\label{tab:32-8-40_best-worst}
\centerline{
\resizebox{1.0\textwidth}{!}{%
\begin{tabular}{clgglgglgglgglgg}
    \toprule
    & \multicolumn{3}{l}{$\mathsf{Precision}$} & \multicolumn{3}{l}{$\mathsf{Recall}$} & \multicolumn{3}{l}{$\mathsf{Jaccard}$} & \multicolumn{3}{l}{$\mathsf{F1}$} & \multicolumn{2}{l}{$\mathsf{Accuracy}$}  \\
          & $M$             & $SD_V$    & $SD_P$        &$M$             & $SD_V$  & $SD_P$            & $M$             & $SD_V$  & $SD_P$         & $M$             & $SD_V$        & $SD_P$       & $M$             & $SD_V$              \\
    \midrule
Run 2 & .843      & .072   & .060  & .814     & .058     & .104   & .699    & .090     & .099  & .804   & .072     & .078   & .867  & .068     \\
Run 4 & .831     & .070    & .068  & .800     & .071     & .111   & .682    & .091    &  .108    & .790  & .073    & .088    & .860  & .065     \\      
    \bottomrule
\end{tabular}
}
}
\end{table}

Finally, using the same fixed set of hyperparameters, we trained the model on the combined set of 32 training videos and 8 validation videos. Here, we performed five experimental runs as before. The results on the 40 test videos are reported in Table~\ref{tab:40-40}. As expected, training on eight additional videos has a considerable effect on the test results. For example, mean $\mathsf{Accuracy}$ increases from 0.865 to 0.872 and mean $\mathsf{Jaccard}$ increases from 0.692 to 0.707. 

\begin{table}[htb]
\caption{Results on the 40:40 data split. For each metric, we report the mean~$M$, standard deviation over videos~$SD_V$, standard deviation over phases~$SD_P$, if applicable, and standard deviation over runs~$SD_R$. $\mathsf{Strategy~B}$ is used for handling undefined values.}
\label{tab:40-40}
    \begin{subtable}[h]{1\textwidth}
    \begin{tabular}{lggglggglggg}
      \toprule
     \multicolumn{4}{l}{$\mathsf{Precision}$} & \multicolumn{4}{l}{$\mathsf{Recall}$} & \multicolumn{4}{l}{$\mathsf{Jaccard}$} \\
             $M$             & $SD_V$    & $SD_P$    & $SD_R$     &$M$             & $SD_V$  & $SD_P$         & $SD_R$    & $M$             & $SD_V$  & $SD_P$      & $SD_R$            \\
        \midrule
        .854     & .067   & .045  & .003    & .811     & .063   & .100     & .007    &  .707   & .091  &  .098   & .007    \\
        
        \bottomrule
    \end{tabular}
    \end{subtable}
    
    \medskip
    \begin{subtable}[h]{1\textwidth}
    \begin{tabular}{lggglgglgg}
      \toprule
     \multicolumn{4}{l}{$\mathsf{F1}$} & \multicolumn{3}{l}{$\mathbf{Macro\,F1}$} & \multicolumn{3}{l}{$\mathsf{Accuracy}$}  \\
        $M$             & $SD_V$        & $SD_P$      & $SD_R$   & $M$             & $SD_V$       & $SD_R$  & $M$             & $SD_V$       & $SD_R$        \\
        \midrule
         .810    & .070 & .077  & .005 & .832 & .060 & .003 & .872 & .062  & .004 \\
        \bottomrule
    \end{tabular}
    \end{subtable}
\end{table}


\section{State of the art on the Cholec80 benchmark}
\label{sec:sota}

In the following tables, we summarize previously published results on the Cholec80 data set. All methods that are referenced in the tables are briefly described in Appendix~\ref{sec:methods}. We consider only results for \emph{online} phase recognition and report the results without additional post-processing steps (section~\ref{sec:post-processing}).  

We made an effort to arrange the reported results in consistent tables, meaning that all results in one table were obtained on the same data split and computed in the same way. For this reason, we present relaxed evaluation results on the 32:8:40 (or 40:40) data split in Table~\ref{tab:sota-40-40-relaxed} and regular evaluation results on the same data split in Table~\ref{tab:sota-40-40}.
Still, many evaluation details were not reported in the papers, so they could not be considered. These details regard, for example, how and over which source of variation the standard deviation was calculated (section~\ref{sec:issue-std}), which variant of the mean F1 score was computed (section~\ref{sec:issue-macro-f1}), and how missing phases in video annotations were handled (section~\ref{sec:undefined}).

\afterpage{%
\clearpage
\begin{landscape}
\begin{table}
\caption{Relaxed results reported on the 32:8:40 data split. \emph{Seq. length} indicates the length of the sequences on which the temporal model components are trained. Here, $T$ indicates that the temporal model is trained on complete videos.  
\emph{Trained on val.} indicates whether the eight validation videos are included in the training data. \emph{Exp. runs} indicates the number of experimental runs. For the \colorbox{tabcolor1!70}{highlighted studies}, it is not completely clear whether or not the results were computed under relaxed conditions. Unfortunately, we did not receive a response by the corresponding authors regarding this question.}
\addcontentsline{toc}{subsection}{Relaxed results reported on the 32:8:40 data split}
\label{tab:sota-40-40-relaxed}
\centering
\begin{adjustbox}{width={1.35\textwidth},keepaspectratio}
\begin{tabular}{lm{3.0cm}lccccclglglglg}
\toprule
\multirow{2}{*}{Model}          & \multirow{2}{*}{\makecell[tl]{Results\\published in}} & \multirow{2}{*}{\makecell[tl]{Seq.\\length}} & \multirow{2}{*}{\makecell[tl]{Trained\\on val. ?}} & \multirow{2}{*}{\makecell[tl]{Exp.\\runs}} & \multirow{2}{*}{\makecell[tl]{Extra\\labels?}} & \multicolumn{2}{l}{Public code?} & \multicolumn{2}{l}{$\EuScript{R}\text{-}\mathsf{Precision}$} & \multicolumn{2}{l}{$\EuScript{R}\text{-}\mathsf{Recall}$} & \multicolumn{2}{l}{$\EuScript{R}\text{-}\mathsf{Jaccard}$} & \multicolumn{2}{l}{$\EuScript{R}\text{-}\mathsf{Accuracy}$} \\
    &       &                                &                                          &                         &                                       & Train.         & Eval.         & $M_P$                        & $SD_P$                        & $M_P$                       & $SD_P$                      & $M_P$                       & $SD_P$                       & $M$                         & $SD_V$                        \\
\midrule
\multirow{4}{*}{\makecell[cl]{SV-RCNet\\\citep{jin2017sv}}}     &   \citet{jin2017sv}         & $2\,\mathrm{s}$   & \cmark                           &    --              & \multirow{5}{*}{\xmark}                                   & \href{https://github.com/YuemingJin/SV-RCNet}{\cmark}           & \href{https://github.com/YuemingJin/SV-RCNet/tree/master/surgicalVideo/evaluation_Cholec80}{\cmark}             & .807                         & .070                          & .835                        & .075                        &                             &                              & .853                        & .073                          \\
            & \citet{jin2020multi}      & $10\,\mathrm{s}$          & \xmark                                   & --                           &                                          &    \href{https://github.com/YuemingJin/MTRCNet-CL}{\cmark}              &     \xmark               & .829                         & .059                          & .845                        & .080                        &                             &                              & .864                        & .073                          \\
            & \hyperref[tab:32-8-40_relaxed]{this paper}         & $8\,\mathrm{s}$                  & \xmark                                   & 5                       &                                          &      tbd            &    \href{https://gitlab.com/nct_tso_public/phasemetrics}{\cmark}                & .873                         & .043                          & .840                        & .100                        & .726                        & .093                         & .874                        & .068                          \\
     & \citet{rivoir2022pitfalls}   & $64\,\mathrm{s}$       & \multirow{2}{*}{\xmark}                                   & \multirow{2}{*}{3}                        &                                          & \multirow{2}{*}{\href{https://gitlab.com/nct_tso_public/pitfalls_bn}{\cmark}}           & \multirow{2}{*}{\href{https://gitlab.com/nct_tso_public/pitfalls_bn/-/blob/main/train_scripts/util_train.py}{\cmark}}             & .894                         & .070                          & .898                        & .064                        & .798                        & .092                         & .913                        & .069                          \\
w/ ConvNeXt &  \citet{rivoir2022pitfalls} & $256\,\mathrm{s}$           &                                    &                            &                                    &            &             & .910                         & .050                          & .908                        & .095                        & .829                        & .101                         & .935                        & .065 \\
\midrule
\makecell[cl]{MTRCNet-CL\\\citep{jin2020multi}}   & \multirow{2}{*}{\citet{jin2020multi}}        & \multirow{2}{*}{$10\,\mathrm{s}$}          & \multirow{2}{*}{\xmark}                                   & \multirow{2}{*}{--}                                & \multirow{2}{*}{tools}                                    & \multirow{2}{*}{\href{https://github.com/YuemingJin/MTRCNet-CL}{\cmark}}           & \multirow{2}{*}{\xmark}             & .869           & .043                              & .880                        & .069                        &                             &                              & .892                        & .076                          \\
MTRCNet   &         &           &                     &                 &                                     &           &            & .850           & .041                              & .851                       & .071                        &                             &                              & .873                        & .074                          \\
\midrule
\makecell[cl]{TMRNet\\\citep{jin2021temporal}}  & \multirow{2}{*}{\citet{jin2021temporal}}    & \multirow{2}{*}{\makecell[cl]{$10\,\mathrm{s}+ 30\,\mathrm{s}$\\\small (memory bank)}}            & \multirow{2}{*}{\cmark}                                   & 5                           & \multirow{2}{*}{\xmark}                                   & \multirow{2}{*}{\href{https://github.com/YuemingJin/TMRNet}{\cmark}}           & \multirow{2}{*}{\href{https://github.com/YuemingJin/TMRNet/tree/main/code/eval/result/matlab-eval}{\cmark}}             & .897                         & .035                          & .895                        & .048                        & .789                        & .058                         & .892                        & .094                          \\
w/ ResNeST               &               &                                &           & --                                 &                                          &                  &                    & .903                         & .033                          & .895                        & .050                        & .791                        & .057                         & .901                        & .076                          \\
\midrule
\makecell[cl]{SSM-LSTM\\\citep{ban2021aggregating}} & \citet{ban2021aggregating} & \makecell[cl]{$8\,\mathrm{s} ~+ $\\\small SSM feature}                            & \cmark                                             & --                                             & \xmark                                         & \xmark          & \xmark         & .870               &                     & .830              &           & &        & .900              &                     \\
\midrule
\rowcolor{tabcolor1!70} \cellcolor{tabcolor1!70} \colorbox{tabcolor1!70}{\makecell[cl]{CBAM-ResNet +\\IndyLSTM +\\Non-Local Block\\\citet{shi2022attention}}}        & \citet{shi2022attention}    &  ?        & \cmark                                   & --                          & \xmark                                   & \xmark           & \xmark             & .878                         &                               & .895                        &                             &                             &                              & .898                        &                             \\ 
\midrule
\rowcolor{tabcolor1!70} \cellcolor{tabcolor1!70} \colorbox{tabcolor1!70}{\makecell[cl]{Non-local 3D~CNN \\+ RLO \\\citep{zhang2021real}}} & \citet{zhang2021real}            &   $160\,\mathrm{s}$                            & \cmark                                   & --                        & \xmark                                   & \xmark    & \xmark & .915                         &                       & .906                        &                         &                       &                         & .905                        &  \\
\midrule
\makecell[cl]{OHFM\\\citep{yi2019hard}}                                    &     \citet{yi2019hard}                 & $T$  & \cmark                                   & 3                               & \xmark                                   & \href{https://github.com/ChinaYi/miccai19}{\cmark} & \xmark     &                              &                               &                             &                             & .670                        & .133                         & .873                        & .057                          \\
\midrule
\makecell[cl]{TeCNO$^{\mathlarger\ast}$\\\citep{czempiel2020tecno}}          & \citet{gao2021trans}   &      $T$         & \cmark                                   & --                       & \xmark                                   & \href{https://github.com/xjgaocs/Trans-SVNet}{\cmark}           & \xmark             & .865                         & .070                          & .876                        & .067                        & .751                        & .069                         & .886                        & .078                          \\
\midrule
\makecell[cl]{Trans-SVNet$^{\mathlarger\ast}$\\\citep{gao2021trans}}         & \citet{gao2021trans}        &   $T$      & \cmark                                   & --              & \xmark                                   & \href{https://github.com/xjgaocs/Trans-SVNet}{\cmark}           & \xmark             & .907                         & .050                          & .888                        & .074                        & .793                        & .066                         & .903                        & .071      \\
\midrule
\makecell[cl]{SAHC$^{\mathlarger\ast\mathlarger\ast}$\\\citep{ding2022exploring}} & \citet{ding2022exploring}             &    $T$                            & ?                                   & --                               & \xmark                                   & \href{https://github.com/xmed-lab/SAHC}{\cmark}    & \href{https://github.com/xmed-lab/SAHC}{\cmark} & .903                         & .064                          & .900                        & .064                        & .812                        & .055                         & .918                        & .081 \\
\midrule
\makecell[cl]{Not E2E\\\citep{yi2022not}} & \citet{yi2022not}                     &      $T$                          & \cmark                                   & --                              & \xmark                                   & \href{https://github.com/ChinaYi/NETE}{\cmark}     & \href{https://github.com/ChinaYi/NETE}{\cmark} &                              &                               & .870                        & .073                        & .771                        & .115                         & .920                        & .053   \\
\bottomrule 
\multicolumn{16}{l}{\footnotesize $^{\mathlarger\ast}$ In the code provided by \citet{gao2021trans}, eight of the test videos are used as validation data for model selection.}\\
\multicolumn{16}{l}{\footnotesize $^{\mathlarger\ast\mathlarger\ast}$ SAHC uses acausal downsampling and unmasked attention operations.}\\
\end{tabular}

\end{adjustbox}
\end{table}

\begin{table}[t]
\caption{Results reported on the 32:8:40 data split. \emph{Trained on val.} indicates whether the eight validation videos are included in the training data. \emph{Exp. runs} indicates the number of experimental runs. $\mathit{F1}$ was computed as harmonic mean of mean precision and mean recall.
For the \colorbox{tabcolor1!70}{highlighted studies}, it is not completely clear whether or not the results were computed under relaxed conditions. Unfortunately, we did not receive a response by the corresponding authors regarding this question.}
\addcontentsline{toc}{subsection}{Regular results reported on the 32:8:40 data split}
\label{tab:sota-40-40}
\centerline{
\centering
\begin{adjustbox}{width={1.45\textwidth},keepaspectratio}
\begin{tabular}{m{3.2cm}m{3.05cm}ccccclgglgglglgglg}
\toprule
\multirow{2}{*}{Model}                             & \multirow{2}{*}{\makecell[tl]{Results\\published in}} &  \multirow{2}{*}{\makecell[tc]{Trained\\on val?}} & \multirow{2}{*}{\makecell[tl]{Exp.\\runs}}  & \multirow{2}{*}{\makecell[tc]{Extra\\labels?}} & \multicolumn{2}{l}{Public code?} & \multicolumn{3}{l}{$\mathsf{Precision}$} & \multicolumn{3}{l}{$\mathsf{Recall}$} & \multicolumn{2}{l}{$\mathsf{Jaccard}$} & \multicolumn{3}{l}{$\mathsf{Accuracy}$} & &  \\
                               &                    &                                                                                         &                                            &                                                 & Train.          & Eval.          & $M$              & $SD_P$ & $SD_R$             & $M$             & $SD_P$  & $SD_R$ & $M$             & $SD_P$          & $M$               & $SD_V$  &  $SD_R$ & $\mathit{F1}$  & $SD_R$            \\
\midrule
\multirow{2}{*}{\makecell[cl]{SV-RCNet\\\citep{jin2017sv}}}   & \citet{zou2022arst}          & ?                                & --                                                            & \multirow{2}{*}{\xmark}                                         & \href{https://github.com/YuemingJin/SV-RCNet}{\cmark}          & \xmark         & .797            &   .081            &         & .792             &  .063  &  &  .655   &   .095         &  .850           &    .068   &   & .794       \\
& \hyperref[tab:results_32-8-40_1]{this paper}         & \xmark                                & 5                                           &                                         & tbd          & \href{https://gitlab.com/nct_tso_public/phasemetrics}{\cmark}         & .839            &   .057               &  .006    & .805             &  .107  & .006 &  .692   &   .103         &  .865           &    .066  & .003      &    .822   & .005 \\
\rowcolor{tabcolor1!70} w/ Swin  & \citet{pan2022temporal}        & ?                                        & --                             & \xmark                                   & \xmark           & \xmark             & .824                         & .065          &                & .778                        & .088        &                &                             &                              & .859                        & .070            &            & .800 & \\
\midrule
\multirow{3}{*}{\makecell[cl]{TeCNO\\\citep{czempiel2020tecno}}}   & \citet{zou2022arst}          & ?                                & --                                                     & \multirow{3}{*}{\xmark}                                         & \multirow{3}{*}{\href{https://github.com/tobiascz/TeCNO}{\cmark}}         & \xmark         & .827             &    .090              &      & .828               &   .065 &  & .696   &     .109          &  .872           &   .077   &     &   .827   \\
  &\cellcolor{tabcolor1!70}\citet{chen2022spatio}          & \cellcolor{tabcolor1!70}\cmark                                & \cellcolor{tabcolor1!70}--                              &                                          &           & \cellcolor{tabcolor1!70}\xmark         & \cellcolor{tabcolor1!70}.821            & \cellcolor{tabcolor1!70}                 & \cellcolor{tabcolor1!70}      & \cellcolor{tabcolor1!70}.858       &  \cellcolor{tabcolor1!70}     &  \cellcolor{tabcolor1!70}   & \cellcolor{tabcolor1!70}.723 & \cellcolor{tabcolor1!70}              & \cellcolor{tabcolor1!70}.900           & \cellcolor{tabcolor1!70}   &  \cellcolor{tabcolor1!70}   & \cellcolor{tabcolor1!70}.839    & \cellcolor{tabcolor1!70} \\
  &  \citet{czempiel2022surgical}           & \cmark                                & 3                         &                                          &           & \xmark         &             &         &         &      &               &     &   &               &  .874         &  & .014       &    .825$^{\mathlarger\ast}$   & .018 \\
 w/ Swin    &  \multirow{2}{*}{\citet{czempiel2022surgical}}           & \multirow{2}{*}{\cmark}                                & \multirow{2}{*}{3}                                                         &      \multirow{2}{*}{\xmark}                                    &   \multirow{2}{*}{\xmark}         & \multirow{2}{*}{\xmark}         &             &       &         &         &               &     &   &               &  .876           &   & .007     &    .817$^{\mathlarger\ast}$ & .016   \\
 w/ X3D      &            &            &              &      &                                           &                                                   &           &         &             &                        &               &     &   &               &  .858           &     & .014   &    .804$^{\mathlarger\ast}$ & .013    \\
\midrule
SENet50 + TCN & \citet{kadkhodamohammadi2022patg}                      & \multirow{3}{*}{\cmark}                    & \multirow{3}{*}{--}                                               & \multirow{3}{*}{\xmark}                                         & \multirow{3}{*}{\xmark}          & \multirow{3}{*}{\xmark}         & .831             &               &         & .821          &    &     &&         &  .883          &    &      &  \makecell[cl]{.826\\.801$^{\mathlarger{\ast\ast}}$ }     \\
SENet50 + LSTM &                       &                   &                                               &                                          &           &        & .830             &         &               & .830     &        &   &&           & .894           &     &    &   \makecell[cl]{.830\\.807$^{\mathlarger{\ast\ast}}$}      \\
\midrule
ResNet + 2-layer GRU   &   \multirow{3}{*}{\citet{czempiel2022surgical}}       & \multirow{3}{*}{\cmark}                                & \multirow{3}{*}{3}                                                           & \multirow{3}{*}{\xmark}                                         & \multirow{3}{*}{\xmark}          & \multirow{3}{*}{\xmark}         &            &          &          &       &        &     &    &             &  .856           &     & .024 &   .767$^{\mathlarger\ast}$   & .017    \\
w/ Swin   &           &                                                              &                    &                 &                &         &           &         &            &                    &               &     &    &             & .877           &   & .022   &   .807$^{\mathlarger\ast}$  & .012      \\
w/ X3D   &           &                                 &                                         &                &             &            &           &         &            &                    &               &     &    &             & .855           &   & .017   &   .790$^{\mathlarger\ast}$     & .014  \\
\midrule
I3D + 4-layer GRU   &   \multirow{4}{*}{\citet{he2022empirical}}       & \multirow{4}{*}{\cmark}                                & \multirow{4}{*}{--}                                                     & \multirow{4}{*}{\xmark}                                         & \multirow{4}{*}{\xmark}          & \multirow{4}{*}{\xmark}         &  .802        &         & .002           &   .806            &     & .020 &    &             &  .883           &      & .010 &     .804     \\
w/ SlowFast   &           &                                                                 &                    &                                          &           &         &   .831         &       & .021             &   .823            &     &  .012 &    &             & .905          &      & .005 & .827    \\
w/ TimeSformer   &           &                                                          &                    &                                          &           &         &  .861          &         & .011           &      .832         &     &  .018 &    &             & .904           &    & .005   &  .846      \\
w/ Video Swin   &           &                                                         &                    &                                          &           &         &      .851      &        & .017            &      .856         &   & .005  &    &             & .909           &     & .000 &    .853     \\
\midrule
\multirow{2}{*}{\makecell[cl]{TransSV-Net\\\citep{gao2021trans}}}   & \citet{zou2022arst}    & ?                                      & --                                                        & \multirow{2}{*}{\xmark}                                         & \multirow{2}{*}{\href{https://github.com/xjgaocs/Trans-SVNet}{\cmark}}         & \xmark         & .850            &     .076        &            & .862            &   .064    & &  .738  &   .107            &  .882          &       .080    &   &  .856  \\
   & \cellcolor{tabcolor1!70}\citet{chen2022spatio}    & \cellcolor{tabcolor1!70}\cmark                                      & \cellcolor{tabcolor1!70}--                                        &                                          &           & \cellcolor{tabcolor1!70}\xmark         & \cellcolor{tabcolor1!70}.817            &  \cellcolor{tabcolor1!70}           &  \cellcolor{tabcolor1!70}         & \cellcolor{tabcolor1!70}.875        &    \cellcolor{tabcolor1!70}  &   \cellcolor{tabcolor1!70}  &  \cellcolor{tabcolor1!70}.731 &   \cellcolor{tabcolor1!70}            &  \cellcolor{tabcolor1!70}.896           &  \cellcolor{tabcolor1!70}  & \cellcolor{tabcolor1!70}  &  \cellcolor{tabcolor1!70}.845 & \cellcolor{tabcolor1!70}       \\
\midrule
ARST\newline\citep{zou2022arst}   & \citet{zou2022arst}          & ?                                & --                                                     & \xmark                                         & \xmark          & \xmark         & .853            &    .075    &                & .849              &  .066  &   &  .730  &   .103            &  .882           &    .075 &        &  .851 &  \\
\midrule
OperA\newline\citep{czempiel2021opera}   & \citet[\href{https://miccai2021.org/openaccess/paperlinks/2021/09/01/352-Paper0235.html}{Rebuttal}]{czempiel2021opera}          & \xmark                                & --                                          & tools                                         & \xmark          & \xmark         & .828            &         &           & .845      &        &     &    &             &  .900     &       &      &   .836       \\
\midrule
\rowcolor{tabcolor1!70} Dual Pyramid\newline Transformer\newline\cite{chen2022spatio} & \citet{chen2022spatio}                          & \cmark                & --                                                           & \xmark                                         & \xmark          & \xmark         & .854            &          &              & .863      &        &     &  .754 &               &  .914         &   &      & .858     &     \\
\midrule
PATG\newline\citep{kadkhodamohammadi2022patg}  & \citet{kadkhodamohammadi2022patg}                             & \cmark             & --                                                       & \xmark                                         & \xmark          & \xmark         & .869             &           &             & .840      &        &   & &           &  .914           &      &   &    \makecell[cl]{.854\\.842$^{\mathlarger{\ast\ast}}$}     \\
\bottomrule
\multicolumn{20}{l}{\footnotesize $^{\mathlarger\ast}$ For each run, \citet{czempiel2022surgical} compute the harmonic mean of $\mathsf{Precision}$ and $\mathsf{Recall}$, which are computed by averaging the phase-wise video-wise results first over videos and then over phases.}\\
\multicolumn{20}{l}{\footnotesize $^{\mathlarger{\ast\ast}}$ Numbers reported by \citet{kadkhodamohammadi2022patg}. Since these numbers are smaller than $\mathit{F1}$, they likely refer to the regular $M(\mathsf{F1})$.}\\

\end{tabular}
\end{adjustbox}
}
\end{table}
\end{landscape}
\clearpage
}

We noted that the source code of many recent methods for surgical phase recognition is not publicly available, making it difficult to reproduce results. 
In some cases, only the code for model training and inference was published, while the code for computing evaluation metrics is missing or difficult to find in the code base.
We also found that it was rather uncommon to use a separate validation set or to conduct more than one experimental run.  
In addition, the reported results were often incomplete, missing measures of variability ($SD_V$, $SD_P$, $SD_R$) and the results for $\mathsf{Jaccard}$ and $\mathsf{F1}$. In contrast to the individual $\mathsf{Precision}$ and $\mathsf{Recall}$ scores, these metrics measure both, how accurately \emph{and} how comprehensively, each phase is recognized in each video and therefore are especially meaningful. 
While it is possible to compute an upper bound, $\mathit{F1}$, of the mean $\mathsf{F1}$ based on the mean values reported for $\mathsf{Precision}$ and $\mathsf{Recall}$, 
the differences between $\mathit{F1}$ and $M(\mathsf{F1})$ can be critical, see Table~\ref{tab:sota-40-40}, In this table, we computed $\mathit{F1}$ for all studies. 
For the results reported by \citet{kadkhodamohammadi2022patg}, we can compare this upper bound to the actual mean $\mathsf{F1}$ score.

Notably, recent papers on surgical phase recognition chose to either report relaxed evaluation metrics (Table~\ref{tab:sota-40-40-relaxed}) or regular evaluation metrics (Table~\ref{tab:sota-40-40}), thus creating two branches of benchmark results that are barely comparable to each other.
Also, we found that the authors who reported relaxed evaluation metrics rarely stated this explicitly in their paper. 
Thus, it was necessary to carefully check the evaluation code or to try and contact the authors to obtain clarity regarding this important evaluation detail. 
As a rule of thumb, if the reported precision and recall scores approach or exceed 90\,\% and, at the same time, the accuracy score is not much higher than 90\,\%, it is very likely that these scores were computed in a relaxed manner.
Note, however, that relaxed evaluation metrics should be considered as deprecated because of the issues described in section~\ref{subsec:relaxed}. Most importantly, the spotted implementation error actually renders the popular MATLAB evaluation script invalid.

We could identify three prominent baseline methods for surgical phase recognition, which were also re-evaluated in other papers: an end-to-end CNN-LSTM (SV-RCNet), a MS-TCN trained on frozen visual feature sequences (TeCNO), and Trans-SVNet, which consists of Transformer blocks that are trained on top of a frozen TeCNO model. 
For these methods, both relaxed and regular evaluation results are available since they were recomputed by different researchers. Again, it is interesting to observe the notable difference between  relaxed and  regular numbers. 

Clearly, the evaluation results reported in different papers for the same method and the same approach to metric calculation (relaxed or regular) deviate as well. Besides random noise, reasons for this could be different strategies for model training, such as batch sampling, learning rate scheduling and data augmentation, different strategies for hyperparameter tuning and model selection, or different configurations of the model architecture regarding, for example, the number of layers, the dimension of the latent space or the backbone used for visual feature extraction.
Notably, \citet{rivoir2022pitfalls} showed that the strategy for training CNN-LSTM models is crucial: By training a CNN-LSTM end to end on very long sequences ($\mathrm{256\,s}$), they achieved state-of-the-art results. 

Some studies compared different temporal models (TCN, LSTM, GRU) and backbones for visual feature extraction (ResNet, Swin, and video models such as X3D or Video Swin) \citep{czempiel2022surgical,he2022empirical,kadkhodamohammadi2022patg}.
Here, TCN, LSTM, and 2-layer GRU seem to perform similarly.
Also, there seems to be no critical difference in performance between different modern 2D~CNN and 3D~CNN backbones. 
However, the 4-layer GRU in combination with Transformer-based video models as feature extractors trained by \citet{he2022empirical} achieves state-of-the-art results. Yet, the authors did not include neither a baseline experiment with a simple 2D~CNN backbone such as ResNet-50 nor ablation studies using a GRU with fewer layers. Therefore, it remains somewhat unclear which factors of the model architecture contribute most to the performance improvement. 

In general, we observe two recent trends to achieve state-of-the-art results on the Cholec80 benchmark. Firstly, training end-to-end models with a larger temporal context, as suggested in TMRNet \citep{jin2021temporal}, \citet{zhang2021real}, or \citet{rivoir2022pitfalls}. Secondly, including Transformer blocks (or message passing mechanisms) in temporal models that operate on frozen visual features, often TeCNO features, which has been proposed in TransSV-Net \citep{gao2021trans}, ARST \citep{zou2022arst}, \citet{chen2022spatio}, or PATG \citep{kadkhodamohammadi2022patg}.
Also, most recent methods do \emph{not} depend on annotations of surgical tools to train parts of the model additionally on the tool recognition task. 

For completeness, we present the results reported on the 40:8:32 data split in Table~\ref{tab:sota-40-8-32}. The initial results on Cholec80, which were computed using a 4-fold cross-validation setup on the test set of 40~videos, are presented in Table~\ref{tab:sota-twinanda}. In the case of 4-fold cross-validation, the computed evaluation results are relatively high. Here, for each validation fold, the model is trained on 30 of the test videos, which would remain unseen when evaluating on the 40:40 data split. Therefore, the high numbers could have been achieved due to additional training data. Evidence for this assumption can be seen in Table~\ref{tab:sota-40-8-32}, where \say{EndoNet + LSTM} achieves worse results on the 40:8:32 data split \citep{czempiel2020tecno}. Moreover, while CNN-LSTM models have been proven to work well for surgical phase recognition, using a more modern backbone, \emph{e.g.}, ResNet-50 instead of AlexNet\,/\,EndoNet \citep{czempiel2020tecno}, and training the model end-to-end \citep{rivoir2022pitfalls} has been shown to achieve better results. 

\begin{table}[tb]
\caption{Results reported on the 40:8:32 data split. $\mathit{F1}$ is the harmonic mean of mean precision and mean recall.}
\addcontentsline{toc}{subsection}{Results reported on the 40:8:32 data split}
\label{tab:sota-40-8-32}
\centerline{
\begin{adjustbox}{width={1.15\textwidth},keepaspectratio}
\begin{tabular}{llcccclglglgclg}
\toprule
\multirow{2}{*}{Model}                             & \multirow{2}{*}{\makecell[tl]{Results\\published in}} &  \multirow{2}{*}{\makecell[tl]{Exp.\\runs}}  & \multirow{2}{*}{\makecell[tc]{Extra\\labels?}} & \multicolumn{2}{c}{Public code?} & \multicolumn{2}{l}{$\mathsf{Precision}$} & \multicolumn{2}{l}{$\mathsf{Recall}$} & \multicolumn{2}{l}{$\mathsf{Accuracy}$} &  & \multicolumn{2}{l}{$f\text{-}\mathsf{F1}$}\\
                                                   &                                                       &                                                    &                                                                                         & Train.          & Eval.          & $M$              & $SD_R$              & $M$             & $SD_R$            & $M$               & $SD_R$            & $\mathit{F1}$   & $M$               & $SD_R$ \\
\midrule
\makecell[cl]{MTRCNet\\\citep{jin2020multi}} & \citet{czempiel2020tecno}                                          & 5                                            & tools                                         & \href{https://github.com/YuemingJin/MTRCNet-CL}{\cmark}          & \xmark         & .761             &     .000                  & .780               &    .001           & .828            &        .000      & .770        \\
\midrule
\makecell[cl]{EndoNet + LSTM\\\citep{twinanda2017vision}} & \citet{czempiel2020tecno}                                          & 5                              & tools                                         & \xmark          & \xmark         & .768            &     .026                   & .721               &    .006           &  .809            &        .002    &  .744         \\
\midrule
\makecell[cl]{ResNet + LSTM\\\small trained separately} & \citet{czempiel2020tecno}                                          & 5                                              & tools                                         & \xmark          & \xmark         & .805             &     .016                   & .799              &    .018           &  .866           &       .010     & .802    \\
\midrule
\makecell[cl]{TeCNO\\\citep{czempiel2020tecno}} & \citet{czempiel2020tecno}                                          & 5                                     & tools                                         & \href{https://github.com/tobiascz/TeCNO}{\cmark}          & \href{https://github.com/tobiascz/TeCNO}{(\cmark)}         & .816              &     .004                 & .852               &   .011                & .886             &        .003      & .834        \\
\midrule 
\makecell[cl]{ResNet + LSTM\\\small trained end to end,\\\small sequence length $64\,\mathrm{s}$} & \citet{rivoir2022pitfalls}                                          & 3                                              & --                                         & \href{https://gitlab.com/nct_tso_public/pitfalls_bn}{\cmark}          & \href{https://gitlab.com/nct_tso_public/pitfalls_bn/-/blob/main/train_scripts/util_train.py}{\cmark}         &              &                       &              &            &  .901           &       .022     &  & .845 & .014   \\
\midrule
\makecell[cl]{ConvNeXt + LSTM\\\small trained end to end,\\\small sequence length $256\,\mathrm{s}$}  & \citet{rivoir2022pitfalls}                                          & 3                                              & --                                         & \href{https://gitlab.com/nct_tso_public/pitfalls_bn}{\cmark}          & \href{https://gitlab.com/nct_tso_public/pitfalls_bn/-/blob/main/train_scripts/util_train.py}{\cmark}         &              &                       &              &            &  .926  & .001     &  & .871 & .005   \\
\bottomrule
\end{tabular}
\end{adjustbox}
}
\end{table}

\begin{table}[tb]
\caption{Results obtained when training the feature extractor on the first 40 videos and then training and testing the temporal model on the remaining 40 videos in a 4-fold cross-validation experiment. \emph{More train data} indicates whether the first 40~videos are added to the training data when performing cross-validation. }
\label{tab:sota-twinanda}
\centerline{
\begin{adjustbox}{width={1.15\textwidth},keepaspectratio}
\begin{tabular}{llccccclglglg}
\toprule
\multirow{2}{*}{Model}                    & \multirow{2}{*}{\makecell[tl]{Results\\ published in}}   & \multirow{2}{*}{\makecell[tc]{More\\train data?}} & \multirow{2}{*}{\makecell[tl]{Exp.\\runs}} &  \multirow{2}{*}{\makecell[tc]{Extra\\labels?}} & \multicolumn{2}{c}{Public code?} & \multicolumn{2}{l}{$\mathsf{Precision}$} & \multicolumn{2}{l}{$\mathsf{Recall}$} & \multicolumn{2}{l}{$\mathsf{Accuracy}$} \\
                                          &                                         &                                                                        &                          &                                & Train.          & Eval.          & $M_P$         & $SD_P$        & $M_P$       & $SD_P$       & $M$          & $SD_V$        \\
\midrule
\multirow{2}{*}{\makecell[cl]{EndoNet + HMM\\\citep{twinanda2016endonet}}} & \citet{twinanda2016endonet}             & \xmark                                               & 5                                             & \multirow{2}{*}{tools}                          & \xmark          & \xmark         & .737          & .161          & .796        & .079         & .817         & .042          \\
                                          & \citet[ch.\,6.4]{twinanda2017vision} & \cmark                                               & --                                    &                                & \xmark          & \xmark         & .744          & .162          & .776        & .067         & .820         & .046          \\
\midrule
\makecell[cl]{EndoNet + LSTM\\\citep{twinanda2017vision}} & \citet[ch.\,6.4]{twinanda2017vision} & \cmark                                               & --                              & tools                          & \xmark          & \xmark         & .844          & .079          & .847        & .079         & .886         & .096          \\
\midrule
\makecell[cl]{Endo3D\\\citep{chen2018endo3d}}             & \citet{chen2018endo3d}                  & \xmark                                               & --                               & tools                          & \xmark          & \xmark         & .813          &               & .877        &              & .912         &  \\
\bottomrule
\end{tabular}
\end{adjustbox}
}
\end{table}

Finally, three recent methods for surgical phase recognition were evaluated on a 60:20 data split, see Table~\ref{tab:sota-60-20}. However, in this case, it is unclear whether the promising evaluation results were achieved because of an improved method or simply because more videos were available for training. Additionally, due to different experimental setups, the results reported in the three different papers may not even be comparable to each other.

\begin{table}[tb]
\caption{Results reported on the 60:20 data split. }
\label{tab:sota-60-20}
\centerline{
\begin{adjustbox}{width={1.2\textwidth},keepaspectratio}
\begin{tabular}{m{3.25cm}m{3.1cm}m{2.25cm}cm{1.25cm}cclglglglg}
\toprule
\multirow{2}{*}{Model}                             & \multirow{2}{*}{\makecell[tl]{Results\\published in}} & \multirow{2}{*}{\makecell[tl]{Evaluation\\details}} & \multirow{2}{*}{\makecell[tl]{Exp.\\runs}}  & \multirow{2}{*}{\makecell[tl]{Extra\\labels?}} & \multicolumn{2}{c}{Public code?} & \multicolumn{2}{l}{$\mathsf{Precision}$} & \multicolumn{2}{l}{$\mathsf{Recall}$} & \multicolumn{2}{l}{$\mathsf{F1}$} & \multicolumn{2}{l}{$\mathsf{Accuracy}$} \\
                                                   &                                                       &                                                                                     &                          &                                                & Train.          & Eval.          & $M$              & $SD_R$              & $M$             & $SD_R$      & $M$             & $SD_R$      & $M$               & $SD_R$              \\
\midrule
\makecell[cl]{MTRCNet-CL\\\citep{jin2020multi}} & \citet{czempiel2021opera} & 5-fold~CV on train videos                                              & --                                                  & tools                                         & \href{https://github.com/YuemingJin/MTRCNet-CL}{\cmark}          & \xmark         & .793            &      .010              & .827             &     .001           &  $\text{.809}^{\mathlarger\ast}$   &  .010   & .856            &            .002           \\
\midrule
ResNet + LSTM\newline\small trained separately & \citet{czempiel2021opera} & 5-fold~CV on train videos                                             & --                                      & tools                                         & \xmark          & \xmark         & .803             &      .011              & .844              &     .009   & $\text{.823}^{\mathlarger\ast}$  &    .008       & .879            &            .008           \\
\midrule
\multirow{2}{*}{\makecell[cl]{TeCNO\\\citep{czempiel2020tecno}}} & \citet{czempiel2021opera} & 5-fold~CV on train videos                                              & --                                                         & tools                                         & \href{https://github.com/tobiascz/TeCNO}{\cmark}          & \xmark         & .809               &      .008              & .874              &     .006          & $\text{.840}^{\mathlarger\ast}$  &   .006   & .891            &            .008           \\
 & \citet{sanchez2022data} & random 20~test + 60~train videos                                              & 3                           & ~~~\xmark                                         & \xmark          & \xmark         &                  &                    &             &             & .838$^{\mathlarger{\ast\ast}}$    &  .025  & .899             &            .015         \\
 \midrule
Multi-task CNN + TCN \citep{sanchez2022data} & \citet{sanchez2022data} &  random 20 test + 48 train videos                                             & 5                                                 & scene segmentation                                         & \xmark          & \xmark         &                  &                    &             &             & .858$^{\mathlarger{\ast\ast}}$   &  .016   & .895            &            .027         \\
\midrule
Multi-task CNN + TCN \citep{sanchez2022data} & \citet{sanchez2022data} & random 20~test + 60~train videos                                              & 3                              & tools +\newline scene segmentation                                         & \xmark          & \xmark         &                  &                    &             &             & .864$^{\mathlarger{\ast\ast}}$   &    & .924           &                     \\
\midrule
\makecell[cl]{OperA\\\citep{czempiel2021opera}} & \citet{czempiel2021opera} & 5-fold~CV on train videos                                             & --                                              & tools                                         & \xmark          & \xmark         & .822                &      .007              & .869              &     .009          & $\text{.845}^{\mathlarger\ast}$  &  .006    & .913            &            .006           \\
\midrule
PATG\newline\citep{kadkhodamohammadi2022patg} & \citet{kadkhodamohammadi2022patg} &  random 20~test + 60~train videos                                             & 5                         & ~~~\xmark                                         & \xmark          & \xmark         & .898                 &      .008              & .891             &     .007         & .882$^{\mathlarger{\ast\ast}}$  & .002    & .938            &            .004           \\
\bottomrule
\multicolumn{15}{l}{\footnotesize $^{\mathlarger\ast}$ Likely computed like $\mathit{F1}$. ~$^{\mathlarger{\ast\ast}}$ Likely computed like $M(\mathsf{F1})$.}\\
\end{tabular}
\end{adjustbox}
}
\end{table}

\section{Conclusion}

In recent years, automatic video-based surgical phase recognition has become an active field of research.
In this paper, we provide a review of previously presented methods for recognizing phases in laparoscopic gallbladder removal (Appendix~\ref{sec:methods}). We summarize the reported results on the Cholec80 benchmark (section~\ref{sec:sota}) and find that a handful of different approaches seem to work similarly well: an end-to-end CNN-LSTM trained with sufficiently large temporal context \citep{rivoir2022pitfalls}, a 4-layer GRU trained on sequences of frozen \emph{Video~Swin} features \citep{he2022empirical}, and some innovative temporal models that integrate attention or message passing mechanisms \citep{gao2021trans,chen2022spatio,kadkhodamohammadi2022patg}. 

However, it seems almost impossible to draw sound conclusions based on the experimental details that were reported in previous studies. In fact, we found that evaluation metrics were (i) obtained on different data splits (section~\ref{sec:datasplits}), (ii) computed in different ways, and (iii) reported incompletely.
In section~\ref{sec:metric-variants}, we present common deviations when calculating evaluation metrics for phase recognition, including (i) different strategies to handle the cases where a phase is missing in the video annotation, (ii) different ways to compute an F1 score, and (iii) calculating the metrics with \say{relaxed boundaries}. 
The effects of these deviations are demonstrated in section~\ref{sec:example}, where we exemplarily evaluate a baseline model to show how differences in computing the evaluation metrics can skew the results.
Unfortunately, it is mostly unknown how exactly evaluation metrics were computed for previous studies. Thus, we conclude that the results of many studies are -- or may be -- actually incomparable. 
Yet, in the past, researchers did not seem to pay much attention to critical evaluation details and included invalid comparisons to prior art in their papers.

Besides inconsistent evaluation metrics, other factors can skew the results as well, for example training details, such as data pre-processing, data augmentation, batch sampling, or learning rate scheduling, and the strategy used for hyperparameter tuning and model selection. Here, one major issue is model tuning and selection on the test set when no separate validation set is used.
While these problems regard the fair evaluation of machine learning models in general, the focus of this paper are the inconsistencies of evaluation metrics used specifically for surgical phase recognition and particularly on the Cholec80 benchmark. 

Regarding the common evaluation metrics for surgical phase recognition (section~\ref{sec:metrics}), we want to note that $\mathsf{Accuracy}$ alone is not well suited to judge classification performance on imbalanced data sets such as Cholec80 (see Fig.~\ref{fig:phase-duration}). With this metric, errors on the shorter phases will be underrepresented.
Therefore, it is common to report the mean $\mathsf{F1}$ or the mean $\mathsf{Jaccard}$ in addition.
We specifically recommend to report not only $M(\mathsf{Precision})$ and $M(\mathsf{Recall})$ because it is not possible to infer $M(\mathsf{F1})$ or $M(\mathsf{Jaccard})$ from these results. Notably, the harmonic mean of $M(\mathsf{Precision})$ and $M(\mathsf{Recall})$ equals $\mathit{F1}$, which typically is considerably larger than $M(\mathsf{F1})$ (see section~\ref{sec:issue-macro-f1}).
Interestingly, \emph{Balanced Accuracy}, which was proposed as a more appropriate evaluation metric on imbalanced data sets, equals $\mathsf{Macro\,Recall}$ \citep{guyon2015design}.  

Still, the \emph{phase-wise} evaluation metrics $\mathsf{Precision}$, $\mathsf{Recall}$, $\mathsf{F1}$, and $\mathsf{Jaccard}$ are not ideal either. With these metrics, errors on the shorter phases are emphasized while errors on the longer phases are weighted less. This also means that ambiguities in phase transitions will impair the evaluation results especially for the shorter phases.   
In this regard, it could be reasonable to relax the evaluation conditions around phase boundaries. However, the commonly used implementation of relaxed metrics is seriously flawed (section~\ref{subsec:relaxed}). 
Further, the video-wise calculation of phase-wise metrics introduces additional problems because  phases 0 and 5 are missing in some Cholec80~videos, meaning that the evaluation results for phases 0 and 5 can, in fact, not be computed for all videos. This is an atypical edge case, which is caused by calculating metrics for each video individually instead of computing metrics over the complete set of all test video frames (section~\ref{sec:frame-wise}). The fact that these edge cases may be handled in different ways (see section~\ref{sec:undefined}) causes further inconsistencies in metric calculation.  
 
For obtaining comparable and consistent evaluation results on the Cholec80 benchmark in future work, we recommend:
\begin{itemize}
    \item Use an \textbf{established data split} with a separate validation set.
    \item To evaluate the final model, \textbf{repeat} the experiment several times, using different random seeds. Summarize the evaluation results over all experimental runs.
    \item \textbf{Publish source code} or describe all details regarding model training and inference. Also state how model hyperparameters were tuned.
    \item \textbf{Publish evaluation code} or describe in detail how the evaluation metrics were computed. Publish the model predictions for each test video and each experimental run, so that deviating evaluation metrics could also be computed in hindsight. An example can be found in our code repository: \url{https://gitlab.com/nct_tso_public/phasemetrics}.
    \item Do not report mean values only, include \textbf{measures of variation} as well.
    \item Besides mean precision and mean recall, also report mean Jaccard or mean F1. \textbf{Do not confuse} $M(\mathsf{F1})$ \textbf{with its upper bound} $\mathit{F1}$ (section~\ref{sec:issue-macro-f1}). 
    \item We identified several \textbf{issues with the relaxed evaluation metrics} (section~\ref{subsec:relaxed}). Do not use them anymore.
    \item Pay attention when compiling comparison tables. Do \textbf{not blindly copy numbers} from previous papers. Check where the results are coming from and whether they were obtained under comparable conditions. A starting point can be the summary tables in section~\ref{sec:sota}. Also, declare whether you adopted results from a previous publication or reproduced the results yourself.    
\end{itemize}

\section{Acknowledgment}

The authors want to thank Deepak Alapatt, Nicolas Padoy, and Lena Maier-Hein for valuable discussions on metrics for surgical phase recognition.

This work was partially funded by the German Research Foundation (DFG, Deutsche Forschungsgemeinschaft) as part of Germany’s Excellence Strategy -- EXC 2050/1 -- Project ID 390696704 -- Cluster of Excellence \say{Centre for Tactile Internet with Human-in-the-Loop}~(CeTI) of Technische Universität Dresden.

\newpage 
\bibliographystyle{abbrvnat}
\bibliography{references} 

\newpage
\appendix
\counterwithin{figure}{section}

\section{Automatic methods for video-based surgical phase recognition}
\label{sec:methods}
Modern methods for video-based surgical phase recognition utilize deep learning models, which are trained on a set~$V_{\mathrm{train}}$ of annotated example videos.
These models typically comprise a visual feature extractor, which learns to compute discriminative representations of individual video frames, and a temporal model, which learns to interpret the sequence of visual features and identify meaningful short-term and long-term temporal patterns for surgical phase recognition. 
Usually, \emph{Convolutional Neural Networks~(CNNs)} or \emph{Vision Transformers~(ViTs)} \citep{dosovitskiy2020image} are employed as feature extractors.

To handle long videos of complete surgical procedures, which can easily exceed an hour, one of two strategies is typically used: two-step training or end-to-end training on truncated sequences. In the first case, visual feature extractor and temporal model are trained separately from each other. More specifically, the feature extractor is trained first. Then, the temporal model is trained on the feature sequences that are obtained by applying the \emph{frozen} visual feature extractor to every frame in the video.
In the latter case, the visual feature extractor and the temporal model are trained jointly on short video sequences, which due to computational limitations usually cover less than one minute of video. 

In the following, we summarize previous approaches for phase recognition that were evaluated on the Cholec80 data set.
Methods that can perform only offline recognition, such as recent works by \citet{zhang2022surgical} and \citet{zhang2022retrieval},  are excluded from this review because they provide limited value for intra-operative surgical assistance.

\subsection{Temporal models trained on frozen visual feature sequences}
 
\citet{twinanda2016endonet} proposed to train \emph{EndoNet}, a CNN based on the \emph{AlexNet} \citep{krizhevsky2017imagenet} architecture, for visual feature extraction. In a second step, they trained a temporal model on the extracted feature sequences, consisting of a \emph{Support Vector Machine~(SVM)} \citep{cortes1995support} to map the EndoNet feature to a $|P|$-dimensional vector and a Hierarchical \emph{Hidden Markov Model~(HMM)} \citep{rabiner1989tutorial} to model temporal dependencies. The authors showed that
(1) training EndoNet in a multi-task manner for both phase recognition and tool presence detection, \emph{i.e.}, surgical tool recognition, improves results and (2) using learned EndoNet features outperforms using hand-crafted features. 
Later, \citet[chapter 6.4]{twinanda2017vision} trained a \emph{Long Short-Term Memory~(LSTM)} \citep{hochreiter1997long} network on frozen EndoNet features and reported improved results. 

Following approaches often used \emph{ResNet-50} \citep{he2016deep}, a deep CNN with residual connections, as visual feature extractor. In many cases, the feature extractor is trained on both the phase recognition and the tool recognition task. However, tool labels may not always be available in other data sets due to the manual annotation effort.
Even so, \citet{sanchez2022data} proposed to train ResNet-50 for phase recognition and surgical scene segmentation, using the annotations of tools and anatomy that the \emph{CholecSeg8k} \citep{hong2020cholecseg8k} data provides for a subset of Cholec80. 

\citet{yi2019hard} noted that recognizing the surgical phase in separate video frames can be difficult due to ambiguous visual appearance and artifacts such as smoke or blur. To better handle these \say{hard} frames, they proposed to (1) identify hard frames in the training data, using a data cleansing strategy \citep{frenay2013classification}, and (2) train ResNet-50 to explicitly recognize hard frames as additional category next to the phases $P$. The ResNet-50 predictions are then further refined by a temporal model, which is trained to substitute the \say{hard} labels with the correct surgical phase. 

\citet{czempiel2020tecno} popularized temporal modeling with \emph{Multi-Stage Temporal Convolutional Networks~(MS-TCNs)} \citep{farha2019ms}. These models consist of several \emph{stages}, where each stage is trained to refine the predictions of the previous stage. In multi-stage TCNs, each stage consists of several layers of dilated 1D~convolutions, where the dilation factor doubles at each layer to increase the size of the receptive field. 
\emph{TeCNO}, proposed by \citet{czempiel2020tecno} for online phase recognition, is a two-stage TCN where the 1D~convolutions are computed in a \say{causal} manner, meaning that the convolution is applied to the current and previous features only and does not access future information. 
 
Since training two-stage models like TeCNO can be difficult when the first stage already generates almost perfect predictions on the training data, \citet{yi2022not} suggest to disturb the first-stage outputs when training the second stage by (1) masking \say{hard} video frames \citep{yi2019hard}, (2) generating the outputs with the first-stage model that did not see the corresponding video during training in a k-fold cross-validation scheme, and by (3) adding Gaussian noise. 

\subsection{Temporal models with Transformer blocks}

Recent approaches integrated \emph{attention mechanisms} \citep{bahdanau2014neural} into temporal models to model long-range temporal dependencies. Simply speaking, an attention mechanism enables each element $q$ in a sequence $Q$ to collect information from all elements in the same or another sequence $V$ by computing an attention-weighted sum, or \emph{context vector}, over the elements $v \in V$. Here, \emph{attention weights} quantify the relevance of each element $v$ given $q$. These weights are computed using a learned attention function, typically scaled dot-product attention \citep{vaswani2017attention}.

Usually, attention mechanisms are contained in Transformer blocks \citep{vaswani2017attention}.
If sequence $V$ is the same sequence as $Q$, attention mechanisms compute \emph{self-attention}, otherwise \emph{cross-attention}. In case of online recognition, where no future information is available, attention needs to be \emph{masked}, meaning that an element at time~$t$ in $Q$ can only attend to, \emph{i.e.}, access information from, elements at time~$t' \leq t$ in $V$. Also, in some cases, attention is calculated in a \emph{local} manner, meaning that an element at time~$t$ can only attend to elements at time~$t'$ with $\vert t - t'\vert \leq \omega $. Here, $\omega$ is the size of the local attention window.

\citet{gao2021trans} proposed \emph{Trans-SVNet}, which adds Transformer blocks to a pre-trained, frozen TeCNO model. First, the TeCNO predictions are passed through a Transformer block with masked self-attention.
 Then, the predictions of the visual feature extractor, a ResNet-50, are passed through a Transformer block with masked cross-attention, attending to the transformed TeCNO predictions. This way, frame-wise predictions can query information from the higher-level transformed TeCNO predictions to obtain refined final phase predictions. 
All attention operations are performed locally with a window size of 30 seconds.

\citet{ding2022exploring} extended the idea to a multi-scale architecture. First, a single-stage TCN is applied to frozen visual features to extract spatiotemporal features at full temporal resolution. Next, the feature sequence is downsampled three times, each time by a factor of 7, using max pooling operations followed by single-stage TCNs. Then, for each of the downsampled sequences, the full-resolution feature sequence is passed through a Transformer block with cross-attention, attending to the downsampled sequence. Thus, the full-resolution feature sequence can query information from higher-level \say{segment} features at multiple temporal scales.  
Note that attention is not masked in this approach. 
The model is trained with supervision at all scales, using downsampled target labels for downsampled features. 

In contrast, \citet{czempiel2021opera} suggested to do without temporal convolutions and simply use a stack of eleven Transformer blocks with masked self-attention for temporal modeling. The proposed approach, \emph{OperA}, is also trained with an \emph{attention regularization loss} to ensure that more attention is being paid to those visual features that correspond to confident phase predictions. 
\citet{he2022empirical} reported difficulties to train a single Transformer block with full self-attention for offline recognition on frozen visual features. 

\citet{kadkhodamohammadi2022patg} proposed to use a 4-layer \emph{Graph Neural Network~(GNN)} \citep{xu2018powerful} with \emph{Principal Neighborhood Aggregation} \citep{corso2020principal} as temporal model. Here, each feature in the frozen visual feature sequence corresponds to one node in the graph and is connected to the nodes that correspond to the 64 previous time steps. 
A ResNet-50 with \emph{Squeeze-and-Excitation} blocks \citep{hu2018squeeze} is employed as visual feature extractor.
Note that a stack of Transformer blocks with local attention can also be seen as an instance of a multi-layer GNN, where information from the neighboring time steps is aggregated using the attention-weighted sum.

\citet{chen2022spatio} proposed to use a two-stage Transformer-based model for temporal modeling, similar to the \emph{ASFormer} by \citet{yi2021asformer}.
Each stage consists of eight modified Transformer blocks with masked local attention, all of which contain a dilated causal 1D~convolution prior to the attention mechanism as \say{local connectivity inductive bias}. 
In the first stage, attention window size and dilation factors are doubled in each block, starting with a size of 1. In the second stage, however, attention window size and dilation factors are halved in each block, starting with a size of 128. Further, the first stage computes self-attention, while the second stage uses cross-attention, thus attending to the first-stage output. 
A Vision Transformer, namely, \emph{Twins-PCPVT} \citep{chu2021twins}, is used as visual feature extractor.

Initially, \citet{vaswani2017attention} introduced a Transformer \emph{encoder-decoder architecture} for sequence-to-sequence translation tasks. Here, the \emph{source} sequence is first processed by the encoder, which is a stack of Transformer blocks with self-attention. Then, the decoder, another stack of Transformer blocks, is used to generate the \emph{target} sequence step by step. To generate the next target token, each decoder block performs two attention operations: first, masked self-attention over its input -- which, at the first block, corresponds to the previously generated target tokens -- and second, full cross-attention over the encoded source sequence. 
With \emph{teacher forcing}  at training time \citep{williams1989learning}, the known target sequence, shifted one token to the right, is used as input to the first decoder block, meaning that all outputs can be generated in parallel.
\citet{zou2022arst} applied a one-block Transformer encoder-decoder with masked local attention to frozen TeCNO features, using an attention window size of five seconds. 
\cite{zhang2022large} trained a Transformer encoder-decoder with six encoder and six decoder blocks 
on frozen C3D features, see section \ref{sec:clip-wise}, using truncated feature segments with a length of 100 as input. However, the achieved evaluation results on the Cholec80 data set were not competitive in this case.
 
\subsection{End-to-end CNN-LSTM or ViT-LSTM models}

\citet{jin2017sv} presented \emph{SV-RCNet}, a combination of a ResNet-50 and an LSTM cell, which is trained end-to-end on short video sequences covering about two seconds. This approach was later extended into a multi-task model, \emph{MTRCNet} \citep{jin2020multi}, which is trained jointly for phase and tool recognition, using longer video sequences of ten seconds. The authors also suggested to further train MTRCNet with a \emph{correlation loss~(CL)}, which ensures that the feature generated in the phase recognition branch can be mapped to the predicted tool probabilities as well. 

Further variants of SV-RCNet were proposed. \citet{pan2022temporal} used a \emph{Swin Transformer} \citep{liu2021swin} as visual feature extractor. \citet{shi2022attention} used a ResNet-50 with \emph{Convolutional Block Attention Modules} \citep{woo2018cbam} as visual feature extractor and an \emph{IndyLSTM} \citep{gonnet2020indylstms} in combination with a \emph{Non-Local Block} \citep{wang2018non} as temporal model. Here, a non-local block is a more generic form of the self-attention mechanism.
 
To model a larger temporal context, \citet{jin2021temporal} proposed to store the frozen SV-RCNet features of the previous 30 time steps in a memory bank \citep{wu2019long}.
After applying a number of parallel temporal convolutions with different kernel sizes to the memory bank, the feature of the current time step is used to query the memory bank by means of a non-local operation. This returns a contextualized representation of the current feature, which is finally used to estimate the current phase. The model is called \emph{TMRNet}. \citet{jin2021temporal} also experimented with a different feature extractor, \emph{ResNeST} \citep{zhang2022resnest}, which could achieve better results.

To model global information from the feature sequence, \citet{ban2021aggregating} proposed \emph{SSM-LSTM}, which continuously stores the LSTM's hidden states as they are being computed. To infer the current phase, an aggregate feature is computed on the history of hidden states and concatenated to the visual feature, which is processed by the LSTM. Here, the aggregate feature is meant to approximate a \say{sufficient statistic feature} of the past.

While \citet[Table VIII]{jin2021temporal} reported difficulties with training CNN-LSTMs on sequences that are longer than 20~seconds, \citet{rivoir2022pitfalls} showed the benefits of increased temporal context by training CNN-LSTMs on sequences of up to 256~seconds. However, training CNN-LSTMs on batches that contain only one long sequence is infeasible when the CNN contains \emph{Batch Normalization~(BN)} layers \citep{ioffe2015batch}, because the requirement that batches are sampled independently and identically distributed (i.i.d.) is violated.
Thus, \citet{rivoir2022pitfalls} used the recent BN-free \emph{ConvNeXt} \citep{liu2022convnet} architecture for visual feature extraction, achieving promising results. 

\subsection{Spatiotemporal visual feature extractors}
\label{sec:clip-wise}

Most methods for phase recognition on Cholec80 used image classification models as feature extractors, which extract visual features from individual frames. A few approaches used video classification models instead, which extract spatiotemporal visual features from short video clips of several, typically 8 -- 64, consecutive frames. \citet{chen2018endo3d} trained a 3D~CNN, \emph{C3D} \citep{tran2015learning}, to extract features from 16-frame clips with a temporal resolution of 2.5 fps. In a second step, they trained a 3-layer LSTM as temporal model. To perform online recognition, clips are sampled every second, using only frames that precede the current time. 

\citet{zhang2021real} proposed to train two 3D~CNNs, both \emph{inflated} \citep{carreira2017quo} ResNet-101 models with non-local blocks \citep{wang2018non}, for surgical phase recognition. One 3D~CNN  operates on 32-frame clips with a temporal resolution of 1 fps. To increase temporal context, the other 3D~CNN operates on 32-frame clips with a temporal resolution of $1/5$ fps, thus covering 160 seconds. To obtain the overall phase prediction, the output logits of both 3D~CNNs are fused by summation. No second-step temporal model is used.   

\citet{czempiel2022surgical} compared three visual feature extractors: (1) a 3D~CNN, \emph{X3D} \citep{feichtenhofer2020x3d}, trained on video clips of 16 sequential frames, (2) a CNN, ResNet-50, and (3) a ViT, Swin. While they found that X3D alone performs better  on Cholec80 than the frame-wise feature extractors, the effect diminishes when additional temporal models are trained on the extracted feature sequences. However, on a larger data set from a different domain, the temporal models also performed best on X3D features.     

\citet{he2022empirical} compared four spatiotemporal visual feature extractors: two 3D~CNNs, \emph{I3D} \citep{carreira2017quo} and \emph{SlowFast} \citep{feichtenhofer2019slowfast}, and two Video Vision Transformers, \emph{TimeSformer} \citep{bertasius2021space} and \emph{Video Swin Transformer} \citep{liu2022video}. They trained these models on 8-frame clips with a temporal resolution of 8 fps, thus covering one second of video. Here, I3D performed a bit worse on Cholec80 than the more recent architectures. In addition, \citet{he2022empirical} trained different temporal models for online and offline recognition on frozen feature sequences. In their experiments, a 4-layer \emph{Gated Recurrent Unit~(GRU)} \citep{cho2014learning} network consistently outperformed a single-stage TCN. 

\subsection{Post-processing strategies}
\label{sec:post-processing}
Some researchers suggested to further refine the model predictions in a post-processing step, typically using hand-crafted rules inferred from knowledge about the Cholec80 workflow (Fig.~\ref{fig:phase-order}) to correct noisy or obviously wrong predictions.

\citet{jin2017sv} proposed the \emph{Prior Knowledge Inference~(PKI)} scheme, which keeps track of the current phase to rectify phase predictions. More specifically, if the current phase is $p$ and the algorithm predicts a phase $q \neq p$ for the next time step, but $q$ cannot follow $p$ immediately in the Cholec80 workflow, \emph{i.e.}, $(p, q) \not \in \EuScript{T}$, then this prediction will be set to $p$. On the other hand, if $(p, q) \in \EuScript{T}$ then the prediction of phase $q$ will be accepted only if it is predicted with a confidence that exceeds $\delta_2$, otherwise, the prediction will be set to $p$ as well. In addition, the current phase $p$ is only updated to the next phase $q$ if $q$ has been predicted for $\delta_1$ continuous time steps. A similar strategy is employed by \citet{pan2022temporal}. However, their method will accept the prediction of the next phase $q$, $(p, q) \in \EuScript{T}$, \emph{iff} $q$ is predicted continuously for the next $\delta$~time steps. Therefore, the method requires access to the next $\delta$ video frames and cannot be computed online. Similarly, the strategy proposed by \citet{zou2022arst} requires access to the next ten video frames in order to check whether the transition to another phase is reliable enough to be accepted. 

\citet{zhang2021real} presented the \emph{progress gate} method, which estimates the current progress to prevent erroneous predictions. 
First, they compute the average duration of the example videos $\bar{T} := \frac{1}{|V_{train}|}\sum_{v \in V_{train}}T(v)$, where $T(v)$ denotes the length of video~$v$.
Then, the progress~$\pi(t)$ at time $t$ is estimated by dividing by the average video duration: $\pi(t) := t / \bar{T}$. 
Furthermore, for each phase $p \in P$, $\pi_{\mathrm{min}}(p) := \mathrm{min} \{ \pi(t) : y(v)_t = p, 0 \leq t < T(v), v \in V_{train} \}$ denotes the lowest progress value at which phase~$p$ occurs in the training data. Analogously, $\pi_{\mathrm{max}}(p) := \mathrm{max} \{ \pi(t) : y(v)_t = p, 0 \leq t < T(v), v \in V_{train} \}$ denotes the highest progress value.
Finally, phase predictions are corrected by enforcing that phase $p$ can only be predicted at time~$t$ if $\pi_{\mathrm{min}}(p) \leq \pi(t) \leq \pi_{\mathrm{max}}(p)$. 


\begin{figure}[htb]
	\centering
    \begin{adjustbox}{clip,trim=0cm 0cm 0cm 1cm,width={0.9\textwidth},keepaspectratio}
        \input{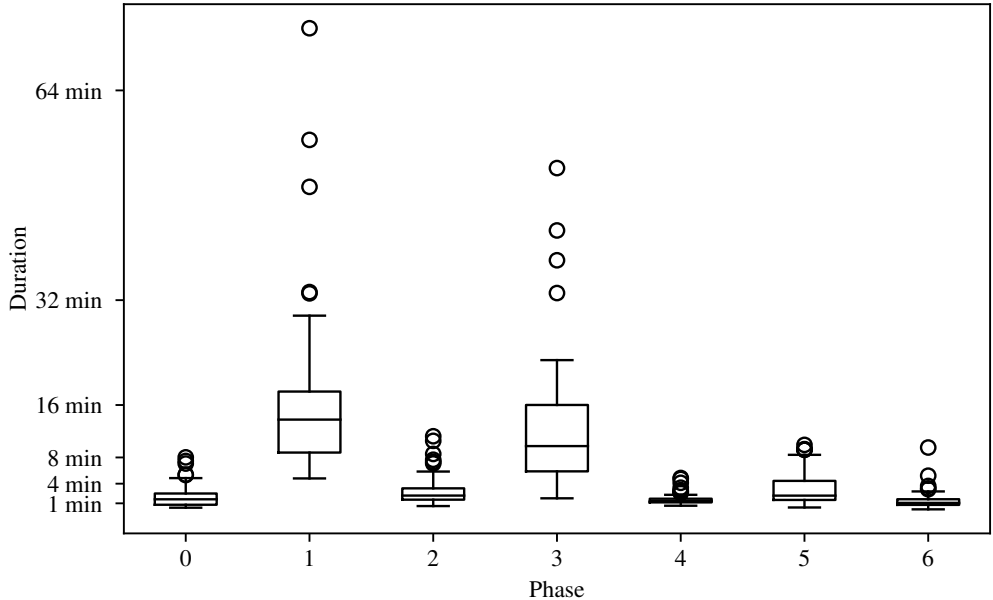}
    \end{adjustbox}
	\caption{Distributions of the duration of each surgical phase in Cholec80, summarized in box plots. } 
	\label{fig:phase-duration}
\end{figure}

\end{document}